\documentclass[10pt,twocolumn,letterpaper]{article}

\usepackage{cvpr}              
\usepackage{multirow}
\definecolor{cvprblue}{rgb}{0.21,0.49,0.74}
\definecolor{lighterblue}{RGB}{220, 235, 250}
\usepackage{colortbl}
\usepackage{graphicx}
\usepackage{amsmath}
\usepackage{amssymb}
\usepackage{booktabs}
\usepackage{algorithm}
\usepackage{algorithmic}
\usepackage[shortlabels]{enumitem}
\usepackage{multirow}
\usepackage{array}       
\usepackage{siunitx}     
\usepackage{xcolor}

\usepackage[pagebackref,breaklinks,colorlinks,allcolors=cvprblue]{hyperref}

\def\paperID{64}
\def\confName{CVPR}
\def\confYear{2026}

\title{The Second Challenge on Cross-Domain Few-Shot Object Detection \\ at NTIRE 2026: Methods and Results}

\author{
\parbox{\textwidth}{
\centering
Xingyu Qiu\textsuperscript{*} \quad
Yuqian Fu\textsuperscript{*} \quad
Jiawei Geng\textsuperscript{*} \quad
Bin Ren\textsuperscript{*} \quad
Jiancheng Pan\textsuperscript{*} \quad
Zongwei Wu\textsuperscript{*} \quad
Hao Tang\textsuperscript{*} \quad
Yanwei Fu\textsuperscript{*} \quad 
Radu Timofte\textsuperscript{*} \quad
Nicu Sebe\textsuperscript{*} \quad
Mohamed Elhoseiny\textsuperscript{*} \quad
Lingyi Hong \quad
Mingxi Cheng \quad
Xingqi He \quad
Runze Li \quad
Xingdong Sheng \quad
Wenqiang Zhang \quad
Jiacong Liu \quad
Shu Luo \quad
Yikai Qin \quad
Yaze Zhao \quad
Yongwei Jiang \quad
Yixiong Zou \quad
Zhe Zhang \quad
Yang Yang \quad
Kaiyu Li \quad
Bowen Fu \quad
Zixuan Jiang \quad
Ke Li \quad
Hui Qiao \quad
Xiangyong Cao \quad
Xuanlong Yu \quad
Youyang Sha \quad
Longfei Liu \quad
Di Yang \quad
Xi Shen \quad
Kyeongryeol Go \quad
Taewoong Jang \quad
Saiprasad Meesiyawar \quad
Ravi Kirasur \quad
Rakshita Kulkarni \quad
Bhoomi Deshpande \quad
Harsh Patil \quad
Uma Mudenagudi \quad
Shuming Hu \quad
Chao Chen \quad
Tao Wang \quad
Wei Zhou \quad
Qi Xu \quad
Zhenzhao Xing \quad
Dandan Zhao \quad
Hanzhe Xia \quad
Dongdong Lu \quad
Zhe Zhang \quad
Jingru Wang \quad
Guangwei Huang \quad
Jiachen Tu \quad
Yaokun Shi \quad
Guoyi Xu \quad
Yaoxin Jiang \quad
Jiajia Liu \quad
Liwei Zhou \quad
Bei Dou \quad
Tao Wu \quad
Zekang Fan \quad
Junjie Liu \quad
Adh\'emar de Senneville \quad
Flavien Armangeon \quad
Mengbers \quad
Yazhe Lyu \quad
Zhimeng Xin \quad
Zijian Zhuang \quad
Hongchun Zhu \quad
Li Wang
}
}

\begin{document}
\maketitle

\begingroup
\renewcommand\thefootnote{*}
\footnotetext{Xingyu Qiu, Yuqian Fu, Jiawei Geng, Bin Ren, Jiancheng Pan, Zongwei Wu, Hao Tang, Yanwei Fu, Radu Timofte, Nicu Sebe, and Mohamed Elhoseiny are the NTIRE 2026 challenge organizers. The other authors are participants in this challenge. \\ 
Appendix~\ref{sec:teams} contains the authors’ team names and affiliations. \\
NTIRE2026 webpage: \href{https://cvlai.net/ntire/2026/}{https://cvlai.net/ntire/2026/}. \\
Challenge Codes: \href{https://github.com/ohMargin/NTIRE2026\_CDFSOD}{https://github.com/ohMargin/NTIRE2026\_CDFSOD}.  
} 
\endgroup

\begin{abstract}
Cross-domain few-shot object detection (CD-FSOD) remains a challenging problem for existing object detectors and few-shot learning approaches, particularly when generalizing across distinct domains. As part of NTIRE 2026, we hosted the second CD-FSOD Challenge to systematically evaluate and promote progress in detecting objects in unseen target domains under limited annotation conditions. The challenge received strong community interest, with 128 registered participants and a total of 696 submissions. Among them, 31 teams actively participated, and 19 teams submitted valid final results. Participants explored a wide range of strategies, introducing innovative methods that push the performance frontier under both open-source and closed-source tracks. This report presents a detailed overview of the NTIRE 2026 CD-FSOD Challenge, including a summary of the submitted approaches and an analysis of the final results across all participating teams.

\end{abstract}  

\section{Introduction}
\label{sec:introduction}

Few-shot object detection (FSOD)~\cite{kohler2023few} aims to enable models to recognize and localize novel object categories from only a handful of labeled examples. Despite notable progress~\cite{shangguan2023identification, qiao2021defrcn, sun2021fsce, wang2020frustratingly, shangguan2024improved, zhang2023detect}, most existing FSOD approaches assume that the training (source) and testing (target) data are drawn from the same domain. However, such an assumption is often violated in real-world scenarios, where models must generalize across substantial domain shifts. For example, detectors trained on natural image datasets such as MS-COCO~\cite{lin2014microsoft} may struggle when applied to domains with significantly different characteristics, such as remote sensing imagery~\cite{gui2024remote,pan2025locate,li2025semantic,liu2025diverse}.

While cross-domain few-shot learning (CD-FSL) has been extensively studied in the context of image classification~\cite{tseng2020cross, guo2020broader, fu2021meta, fu2022me, zhang2022free, li2022cross, fu2023styleadv, zhuo2022tgdm, tang2022learning, zha2023boosting, ren2024sharing, zhuo2024unified, zhuo2026segdp}, its extension to object detection, namely cross-domain few-shot object detection (CD-FSOD), remains relatively underexplored with few exceptions. Typically, CD-ViTO~\cite{fu2024cross} first formally defines this task, and constructs a comprehensive CD-FSOD benchmark which takes COCO as source data and six novel datasets (ArTaxOr, Clipart1k, DIOR, DeepFish, NEU-DET, UODD) as targets.  Following CD-ViTO, newly proposed methods include CDFormer~\cite{meng2025cdformer}, StyleProto~\cite{yang2026styleproto}, ETS~\cite{pan2025enhance}, Domain-RAG~\cite{li2025domain}, LMP~\cite{wang2026learning}, tackling CD-FSOD from various perspectives.

Though with the emergence of few methods, the CD-FSOD task is still brand new and also challenging. Thus, we are motivated to host the challenge to further promote the advances on CD-FSOD task. Specifically, we expand the evaluation to three additional unseen target domains, namely RUOD~\cite{RUOD23}, CARPK~\cite{hsieh2017drone}, and CarDD~\cite{wang2023cardd}, to further assess the generalization capability of CD-FSOD models. Consistent with the findings of CD-ViTO, these datasets exhibit substantial domain discrepancies with respect to the source data, characterized by variations in visual style, inter-class variance (ICV), and ambiguous category boundaries (IB). 
As for the specific task settings, following the 1st CD-FSOD challenge~\cite{fu2025ntire}, we keep two sub settings: \textit{closed-source CD-FSOD} task and \textit{open-source CD-FSOD} task, supporting systematically studying of models.  Particularly, the \textit{closed-source CD-FSOD} refers the one initiall proposed in CD-ViTO protocol, which training data is strictly restricted to the predefined source domain (e.g., MS-COCO); while the more flexible \textit{open-source CD-FSOD} task is proposed for lifting this constraint, allowing participants to exploit additional data sources, prior knowledge, and large-scale foundation models, thereby exploring the upper-bound performance on the target domains.

Formally, organized as part of the 2026 New Trends in Image Restoration and Enhancement (NTIRE 2026) Workshop, which emphasizes robustness under varying conditions, the second CD-FSOD Challenge is introduced to advance research in this area. The challenge consists of two tracks: an open-source CD-FSOD track as the primary track, and a closed-source CD-FSOD track as a supplementary track. In the closed-source setting, MS-COCO is used as the exclusive source domain for training. 
The validation phase follows the protocol of CD-ViTO and includes six predefined target domains. In addition, three newly introduced domains are reserved for the final evaluation in both tracks. Performance is assessed using mean Average Precision (mAP) as the official ranking metric. We anticipate that this challenge will stimulate further progress in CD-FSOD and encourage the development of more robust and generalizable detection approaches.

This challenge is one of the challenges associated with the NTIRE 2026 Workshop~\footnote{\url{https://www.cvlai.net/ntire/2026/}} on: deepfake detection~\cite{ntire26deepfake}, 
high-resolution depth~\cite{ntire26hrdepth},
multi-exposure image fusion~\cite{ntire26raim_fusion}, 
AI flash portrait~\cite{ntire26raim_portrait}, 
professional image quality assessment~\cite{ntire26raim_piqa},
light field super-resolution~\cite{ntire26lightsr},
3D content super-resolution~\cite{ntire263dsr},
bitstream-corrupted video restoration~\cite{ntire26videores},
X-AIGC quality assessment~\cite{ntire26XAIGCqa},
shadow removal~\cite{ntire26shadow},
ambient lighting normalization~\cite{ntire26lightnorm},
controllable Bokeh rendering~\cite{ntire26bokeh},
rip current detection and segmentation~\cite{ntire26ripdetseg},
low light image enhancement~\cite{ntire26llie},
high FPS video frame interpolation~\cite{ntire26highfps},
Night-time dehazing~\cite{ntire26nthaze,ntire26nthaze_rep},
learned ISP with unpaired data~\cite{ntire26isp},
short-form UGC video restoration~\cite{ntire26ugcvideo},
raindrop removal for dual-focused images~\cite{ntire26dual_focus},
image super-resolution (x4)~\cite{ntire26srx4},
photography retouching transfer~\cite{ntire26retouching},
mobile real-word super-resolution~\cite{ntire26rwsr},
remote sensing infrared super-resolution~\cite{ntire26rsirsr},
AI-Generated image detection~\cite{ntire26aigendet},
cross-domain few-shot object detection~\cite{ntire26cdfsod},
financial receipt restoration and reasoning~\cite{ntire26finrec},
real-world face restoration~\cite{ntire26faceres},
reflection removal~\cite{ntire26reflection},
anomaly detection of face enhancement~\cite{ntire26anomalydet},
video saliency prediction~\cite{ntire26videosal},
efficient super-resolution~\cite{ntire26effsr},
3d restoration and reconstruction in adverse conditions~\cite{ntire26realx3d},
image denoising~\cite{ntire26denoising},
blind computational aberration correction~\cite{ntire26aberration},
event-based image deblurring~\cite{ntire26eventblurr},
efficient burst HDR and restoration~\cite{ntire26bursthdr},
low-light enhancement: `twilight cowboy'~\cite{ntire26twilight},
and efficient low light image enhancement~\cite{ntire26effllie}.

\section{NTIRE 2026 CD-FSOD Challenge}
\subsection{Challenge Overview} \label{sec:overview}
\begin{figure*}[h]
\centering	{\includegraphics[width=1.\linewidth]{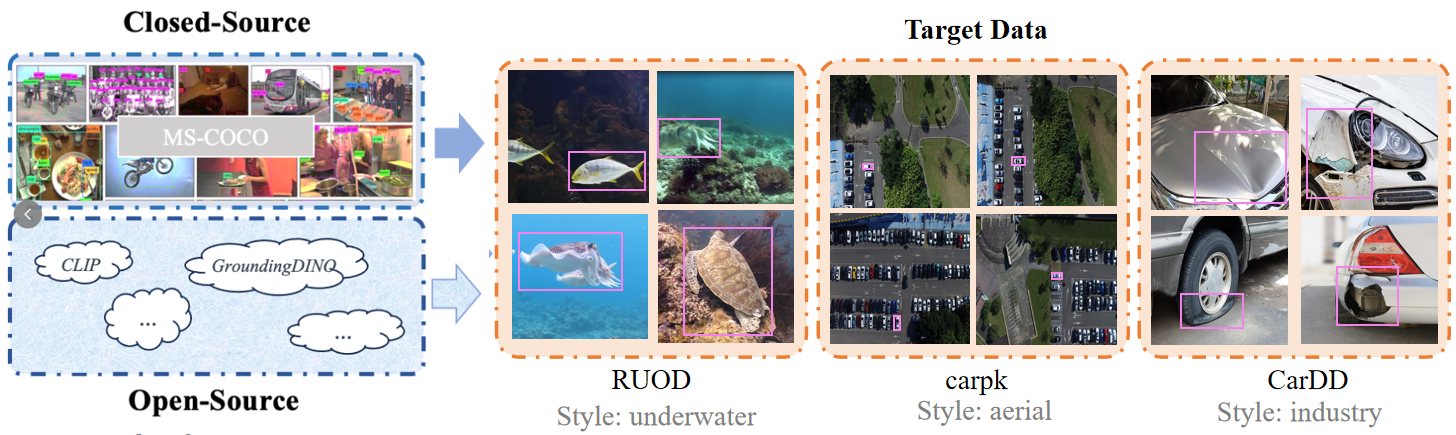}}
\caption{Illustration of the challenge settings, including the closed-source and open-source CD-FSOD tracks. The three newly introduced target datasets used in the final testing phase are also shown.} \label{fig:task}
\end{figure*}

Our challenge aims to advance \textbf{Cross-Domain Few-Shot Object Detection (CD-FSOD)}—detecting objects across domain shifts with limited labeled data. We use six previously published target domains~\cite{fu2024cross} as validation sets and introduce three newly constructed datasets for final testing. In addition to these dataset updates, we propose \textit{open-source CD-FSOD}, allowing participants to freely select source datasets and pre-trained models to improve generalization. Fig.~\ref{fig:task} presents both the predefined closed-source CD-FSOD setting and the open-source CD-FSOD setting, together with the newly introduced target domains.

\subsection{Task Formulations}
\noindent\textbf{Closed-Source CD-FSOD.} Given a source dataset $\mathcal{D}_{S}$ and a novel target dataset $\mathcal{D}_{T}$, the closed-source CD-FSOD track assumes that the source class set $\mathcal{C}_{S}$ and the target class set $\mathcal{C}_{T}$ are completely disjoint, i.e., $\mathcal{C}_{S} \cap \mathcal{C}_{T} = \emptyset$. Additionally, the distributions of the source domain $\mathcal{D}_{S}$ and the target domain $\mathcal{D}_{T}$ are not identical. Participants are required to train models on $\mathcal{D}_{S}$ and test them on $\mathcal{D}_{T}$, where each class in $\mathcal{C}_{T}$ has only a few labeled examples. Usually, $\mathcal{D}_{S}$ is a single dataset, as in CD-ViTO~\cite{fu2024cross}. We refer to this setting as closed-source CD-FSOD to differentiate it from the open-source variant.

\noindent\textbf{Open-Source CD-FSOD.} In contrast to the closed-source setting where training data is strictly limited, the open-source CD-FSOD track is designed to leverage the capabilities of foundation models. Since these models are pretrained on large-scale and diverse datasets, it is practically hard to trace all the knowledge embedded within them.
Hence, we refer to this setting as \textit{open-source}. While the relaxed constraints on source data make it difficult to strictly ensure non-overlapping classes between the source and target data, the track still focuses on addressing the core challenges of domain shift and few-shot object detection. We believe this setting will significantly accelerate the development of CD-FSOD methods for real-world applications.

In this challenge, the open-source CD-FSOD is designated as the main track, with awards presented to the top three teams. The closed-source CD-FSOD serves as the special track, with a single award granted to the top-performing team.

\noindent \textbf{$N$-way $K$-shot Protocol.} We adopt the $N$-way $K$-shot evaluation protocol. For each novel class in the target class set $\mathcal{C}_{T}$, $K$ labeled instances are provided, forming the support set $S$. The remaining unlabeled instances constitute the query set $Q$. Instances contained in the support set $S$ are used to assist the model in recognizing and detecting the objects in $Q$.

\subsection{Challenge Phases and Datasets}
This challenge involves one development stage and one testing stage.  The source data $\mathcal{D}_{S}$ for both stages is the same, i.e., MS-COCO~\cite{lin2014microsoft} for the closed-source track and unlimited data for the open-source track. While the testing data $\mathcal{D}_{T}$ is different. 

\noindent\textbf{Development Stage:} Datasets proposed in the CD-ViTO, including ArTaxOr~\cite{GeirArTaxOr}, Clipart1K~\cite{inoue2018cross}, DIOR~\cite{li2020object}, DeepFish~\cite{saleh2020realistic}, NEU-DET~\cite{song2013noise}, and UODD~\cite{jiang2021underwater} are taken as targets $\mathcal{D}_{T}$ during development stage. 

\noindent\textbf{Testing Stage.} Three previously unseen datasets (RUOD~\cite{RUOD23}, Carpk~\cite{hsieh2017drone}, and CarDD~\cite{wang2023cardd}) are introduced and used as the targets $\mathcal{D}_{T}$ for the final testing phase. Note that the ground truth annotations for these query sets are held exclusively by the challenge organizers.

\subsection{CD-ViTO Baseline Model}
\label{sec:baseline_model}
\begin{figure}[t]
\centering	{\includegraphics[width=1.\linewidth]{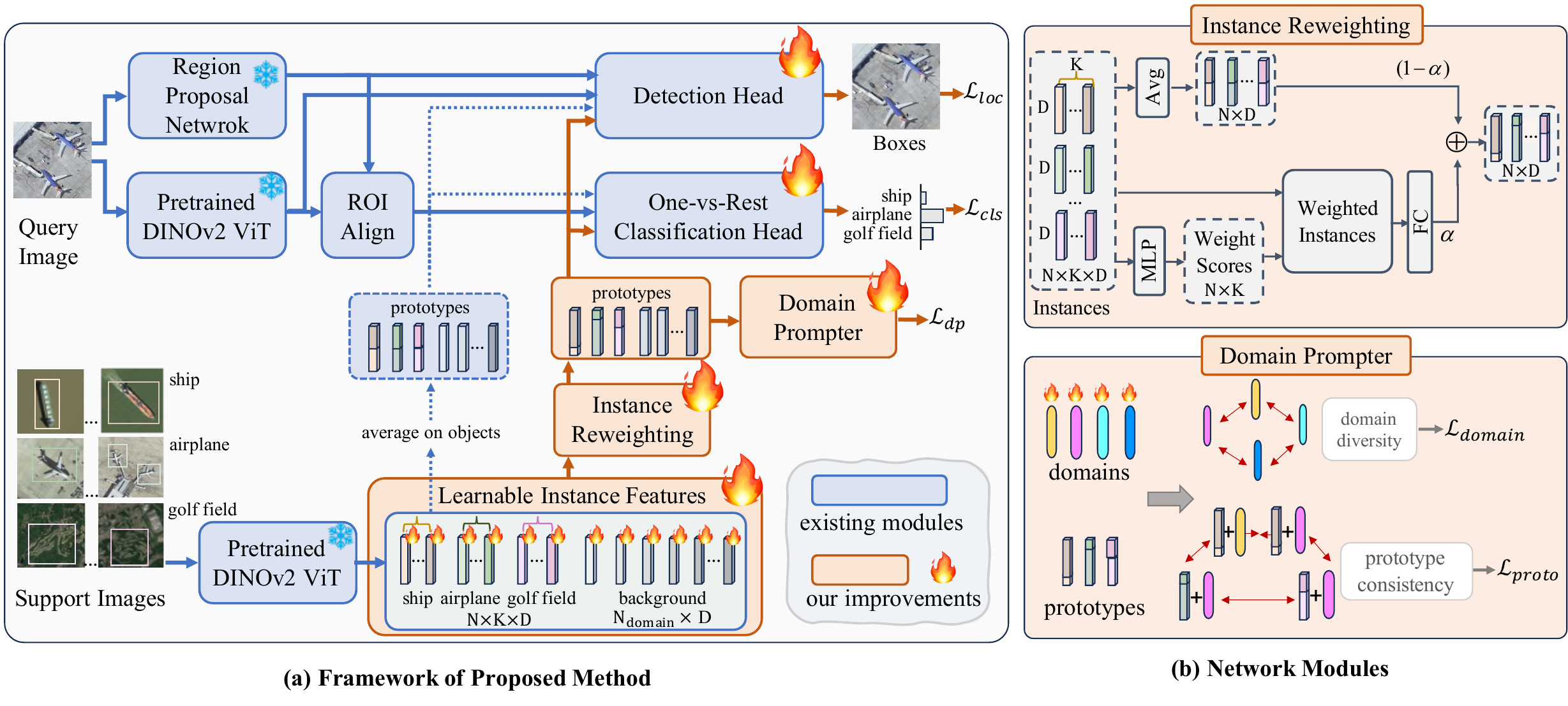}}
\caption{Overall framework of CD-ViTO baseline method. \label{fig:framework-base} 
\vspace{-0.15in}
}
\end{figure}

We take CD-ViTO~\cite{fu2024cross} as the baseline for the closed-source track. Briefly, CD-ViTO is built upon DE-ViT~\cite{zhang2023detect} and fine-tuned using the support set. As in Fig.~\ref{fig:framework-base}, modules in blue are inherited from DE-ViT, while modules in orange are newly proposed. New improvements include 
\textit{learnable instance features}, \textit{instance reweighting}, 
\textit{domain prompter}, and \textit{finetuning}.  

Intuitively, the learnable instance feature module is designed to enhance inter-class variance (ICV) among different target classes by making the initially fixed instance features learnable and optimizing them through supervised few-shot detection tasks on the target support set. The instance reweighting module further improves prototype quality by assigning higher weights to high-quality object instances—e.g., those with minimal indefinable boundary (IB). These weights are learned via a lightweight MLP and fully connected layer, as illustrated in the upper part of Fig.~\ref{fig:framework-base}(b). The domain prompter module introduces learnable domain perturbations to simulate varying domain styles. These perturbations are applied to object prototypes, followed by a prototype consistency loss to ensure that the introduced perturbations do not affect the semantic category of the prototypes. Simultaneously, a domain diversity loss encourages the generated domains to be sufficiently diverse. 
The lower part of Fig.~\ref{fig:framework-base}(b) illustrates this mechanism. By injecting virtual domains and enforcing robustness against the induced perturbations, this strategy enhances the model’s generalization under domain shifts.
Finetuning is applied to the modules highlighted with fire icons in Fig.~\ref{fig:framework-base}.

\subsection{Domain-RAG Baseline Model}
\label{sec:rag_baseline_model}
We take Domain-RAG~\cite{li2025domain}, the current state-of-the-art (SOTA) method, as the baseline for the open-source track. Domain-RAG is a retrieval-guided compositional image generation framework proposed for cross-domain few-shot object detection (CD-FSOD). The key idea is to enhance training samples with domain-consistent synthetic backgrounds without introducing extra supervision or additional model training. 

\begin{figure}[t]
    \centering
    \includegraphics[width=0.95\columnwidth]{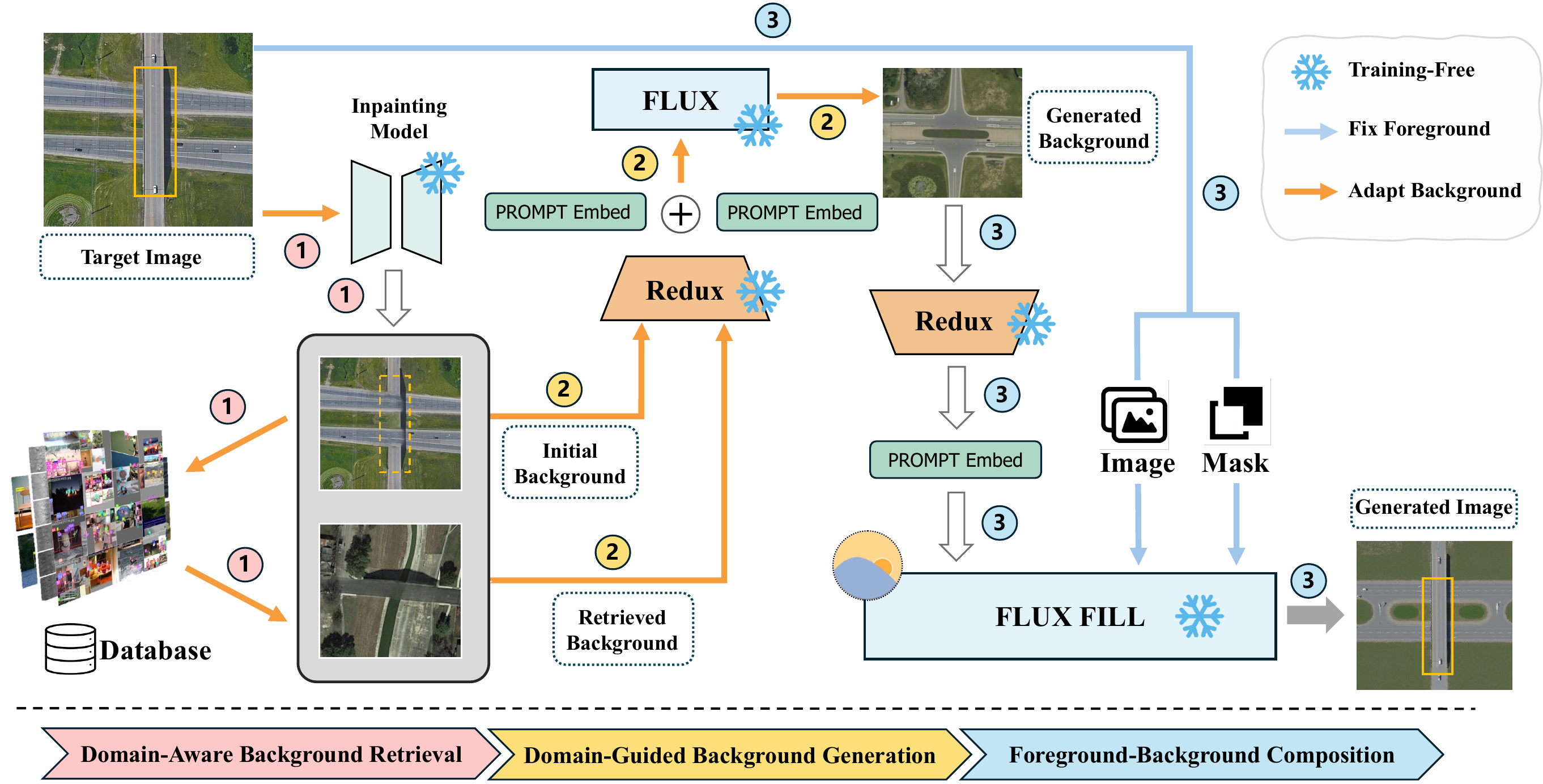}
    \caption{\textbf{Overview of Domain-RAG.}
    Domain-RAG consists of three stages:
    (1) Domain-aware background retrieval,
    (2) Domain-guided background generation, and
    (3) Foreground-background composition.
    The whole pipeline is training-free and follows the principle of \emph{fix foreground, adapt background}.}
    \label{fig:domainrag_framework}
    \vspace{-2mm}
\end{figure}

As shown in Fig.~\ref{fig:domainrag_framework}, Domain-RAG first retrieves domain-relevant background candidates from large-scale image corpora, then generates target-domain-aligned backgrounds conditioned on the retrieved context, and finally composes them with the original foreground objects to form realistic augmented samples. This design preserves foreground semantics while narrowing the domain gap at the background level, thereby improving the detector's robustness and generalization ability in low-shot cross-domain scenarios. Notably, as a data augmentation method, in principle, Domain-RAG could be adapted into any base methods.

\subsection{Evaluation Protocol}
The final score is measured based on the model's performance on the three datasets of the testing stage. For each dataset, we validate the models on three different few-shot settings: 1-shot, 5-shot, and 10-shot. This results in a total of nine mean Average Precision (mAP) scores: \texttt{D1\_1shot}, \texttt{D1\_5shot}, \texttt{D1\_10shot}; \texttt{D2\_1shot}, \texttt{D2\_5shot}, \texttt{D2\_10shot}; and \texttt{D3\_1shot}, \texttt{D3\_5shot}, \texttt{D3\_10shot}. The \texttt{D1, D2, D3} denote the Deep-Fruits, Carpk, and CarDD, respectively. 

The final ranking score is computed as a weighted average $avg()$ of these scores:
\begin{align*}
\textit{Score} =\ & 2*\textit{avg}(\texttt{D1\_1shot}, \texttt{D2\_1shot}, \texttt{D3\_1shot}) \\
&+ 1*\textit{avg}(\texttt{D1\_5shot}, \texttt{D2\_5shot}, \texttt{D3\_5shot}) \\
&+ 1*\textit{avg}(\texttt{D1\_10shot}, \texttt{D2\_10shot}, \texttt{D3\_10shot})
\end{align*}

\noindent\textbf{Rationale for Weighted Scoring.}
We assign a higher weight (×2) to the 1-shot setting for two primary reasons: (1) Performance in the 1-shot scenario is generally lower than in the 5-shot and 10-shot settings due to the limited availability of labeled examples for adaptation; and (2) emphasizing 1-shot performance encourages the development of models that are more robust and effective in extremely low-data conditions.

\begin{table*}[ht]
\centering
\caption{Open-source and closed-source results on CD-FSOD (2026). D1, D2, and D3 represent RUOD, CARPK, and CarDD, respectively. Mean Average Precision (mAP) on 1-shot, 5-shot, and 10-shot are reported. Teams achieving top results are highlighted.}
\label{tab:result}
\resizebox{\linewidth}{!}{
\begin{tabular}{|c|l|c|c|c|c|c|c|c|c|c|c|}
\hline
\multicolumn{12}{|c|}{\cellcolor{blue!25} \textbf{Main Open-Source Track (2026)}} \\
\hline
\textbf{Rank} & \textbf{Team Name} & \textbf{Score} & \textbf{D1\_1shot} & \textbf{D1\_5shot} & \textbf{D1\_10shot} & \textbf{D2\_1shot} & \textbf{D2\_5shot} & \textbf{D2\_10shot} & \textbf{D3\_1shot} & \textbf{D3\_5shot} & \textbf{D3\_10shot} \\
\hline
\cellcolor{blue!15}1 & \cellcolor{blue!15}FDUROILab\_Lenovo & \cellcolor{blue!15}\textbf{217.21} & \cellcolor{blue!15}\textbf{57.04} & \cellcolor{blue!15}\textbf{57.15} & \cellcolor{blue!15}\textbf{58.08} & \cellcolor{blue!15}59.23 & \cellcolor{blue!15}59.23 & \cellcolor{blue!15}59.23 & \cellcolor{blue!15}\textbf{45.23} & \cellcolor{blue!15}46.17 & \cellcolor{blue!15}48.77 \\
\cellcolor{blue!15}2 & \cellcolor{blue!15}CDiscover & \cellcolor{blue!15}192.79 & \cellcolor{blue!15}34.61 & \cellcolor{blue!15}41.14 & \cellcolor{blue!15}42.06 & \cellcolor{blue!15}\textbf{63.26} & \cellcolor{blue!15}\textbf{63.00} & \cellcolor{blue!15}61.29 & \cellcolor{blue!15}39.71 & \cellcolor{blue!15}\textbf{47.43} & \cellcolor{blue!15}48.30 \\
\cellcolor{blue!15}3 & \cellcolor{blue!15}NJUST-KMG & \cellcolor{blue!15}191.38 & \cellcolor{blue!15}35.62 & \cellcolor{blue!15}47.51 & \cellcolor{blue!15}46.22 & \cellcolor{blue!15}60.41 & \cellcolor{blue!15}60.51 & \cellcolor{blue!15}61.12 & \cellcolor{blue!15}40.09 & \cellcolor{blue!15}42.01 & \cellcolor{blue!15}44.54 \\
4 & earth-insights & 190.09 & 38.20 & 44.95 & 46.59 & 58.73 & 62.78 & \textbf{63.63} & 33.95 & 40.10 & \textbf{50.48} \\
5 & Intellindust\_AI\_Lab & 188.05 & 39.61 & 43.05 & 45.25 & 53.42 & 53.60 & 53.29 & 44.86 & 45.82 & 47.37 \\
6 & SAIDA & 161.08 & 30.49 & 39.38 & 37.14 & 56.28 & 56.80 & 55.94 & 30.92 & 27.95 & 30.67 \\
7 & KLETech-CEVI & 159.83 & 22.11 & 23.04 & 21.63 & 61.86 & 60.46 & 60.30 & 32.24 & 39.00 & 42.64 \\
8 & Manifold & 159.41 & 29.31 & 33.91 & 33.40 & 58.26 & 58.26 & 58.26 & 21.78 & 35.09 & 40.60 \\
9 & QiFans & 155.41 & 23.42 & 23.42 & 23.42 & 57.06 & 57.06 & 57.06 & 36.08 & 36.08 & 36.08 \\
10 & AIRCAS\_MILab & 150.61 & 21.30 & 30.82 & 34.14 & 57.11 & 55.35 & 59.66 & 18.36 & 37.06 & 41.23 \\
11 & J\_G\_team & 149.95 & 26.71 & 38.47 & 34.86 & 57.99 & 57.94 & 57.51 & 18.01 & 26.78 & 28.87 \\
12 & NTR & 149.76 & 26.89 & 38.23 & 35.03 & 58.84 & 58.71 & 58.23 & 17.29 & 25.82 & 27.22 \\
13 & WRC & 139.74 & 15.63 & 31.44 & 27.59 & 53.20 & 54.75 & 54.21 & 21.92 & 33.32 & 36.42 \\
14 & NUDT-RSIP & 131.41 & 13.40 & 17.36 & 21.64 & 53.00 & 54.45 & 55.04 & 23.82 & 30.71 & 34.60 \\
15 & French Borelli & 118.05 & 21.25 & 25.89 & 29.29 & 35.87 & 41.14 & 51.55 & 16.10 & 26.93 & 32.91 \\
\hline
\multicolumn{12}{|c|}{\cellcolor{blue!25} \textbf{Special Closed-Source Track (2026)}} \\
\hline
\textbf{Rank} & \textbf{Team Name} & \textbf{Score} & \textbf{D1\_1shot} & \textbf{D1\_5shot} & \textbf{D1\_10shot} & \textbf{D2\_1shot} & \textbf{D2\_5shot} & \textbf{D2\_10shot} & \textbf{D3\_1shot} & \textbf{D3\_5shot} & \textbf{D3\_10shot} \\
\hline
\cellcolor{blue!15}1 & \cellcolor{blue!15}FewShotEverything & \cellcolor{blue!15}\textbf{134.31} & \cellcolor{blue!15}23.02 & \cellcolor{blue!15}29.48 & \cellcolor{blue!15}31.09 & \cellcolor{blue!15}\textbf{41.53} & \cellcolor{blue!15}\textbf{46.65} & \cellcolor{blue!15}\textbf{51.89} & \cellcolor{blue!15}\textbf{21.78} & \cellcolor{blue!15}\textbf{34.82} & \cellcolor{blue!15}\textbf{36.32} \\
2 & Fusion-Few & 108.48 & \textbf{24.48} & \textbf{33.29} & \textbf{33.49} & 27.94 & 27.82 & 27.90 & 15.94 & 31.77 & 34.44 \\
3 & nudt\_0110Dplter & 73.71 & 12.06 & 17.52 & 21.07 & 6.49 & 14.79 & 25.45 & 21.48 & 29.14 & 33.10 \\
4 & freav & 69.82 & 13.31 & 17.44 & 18.76 & 8.41 & 20.26 & 16.55 & 15.92 & 28.65 & 32.54 \\
\hline
\end{tabular}}
\end{table*}

\section{Challenge Results}
Among the 128 registered participants, 15 and 4 teams have participated the final testing stage and submitted their results, codes, and factsheets. Table.~\ref{tab:result} summarizes the results of these methods.  Detailed descriptions of the participants' solutions are provided in Sec.\ref{sec:teams-solution} and Sec.\ref{sec:teams-solution2}, each corresponding to a different track.

\noindent\textbf{Open-Source Track Results.} In the open-source track, nearly all participating teams achieved strong performance with clear improvements over the provided CD-ViTO baseline. This highlights not only the effectiveness of their proposed methods but also the significance of introducing this task setting. As observed, relaxing the strict limitation on the source data offers a substantial advantage in tackling the CD-FSOD task.

Specifically, the teams FDUROILab\_Lenovo, CDiscover, and NJUST-KMG emerged as the top performers in this track, achieving scores of 217.21, 192.79, and 191.38, respectively—significantly outperforming the baseline and the other competing teams in the same track.

\noindent\textbf{Closed-Source Track Results.} The performance achieved by the closed-source track teams is generally lower than that of the open-source track. This is quite understandable considering that the closed-source track enforces stricter constraints. Nevertheless, the participants managed to improve the baseline method clearly.

In particular, the FewShotEverything team stands out with a final score of 134.31, significantly outperforming the other competitors in the Special Closed-Source Track. As shown in Fig.~19, the framework first uses an image generation model to synthesize underwater images from text prompts, and then employs a vision-language model to produce pseudo labels for the generated samples. The strong results suggest that such a pipeline, which combines data generation with automatic annotation, can effectively enrich the training data and improve detection performance under the cross-domain few-shot setting. Other teams in this track also delivered meaningful improvements, demonstrating the promise of closed-source large models for data augmentation.

\section{Main Open-Source Track Methods}
\label{sec:teams-solution}

\subsection{FDUROILab Lenovo}
\subsubsection{Proposed Method}


To significantly enhance the model’s adaptability in complex cross-domain scenarios, the team proposes an efficient fine-tuning strategy tailored for the open-vocabulary detection model. Their approach leverages diverse data augmentation techniques to expand the limited training set and improve the model’s ability to recognize novel objects in the target domain using the provided k-shot annotated samples.

\begin{figure}[h]
    \centering
    \def\figPath{teams/team9_FDUROILab_Lenovo/figure1.pdf}
    \IfFileExists{\figPath}{
        \includegraphics[width=\columnwidth]{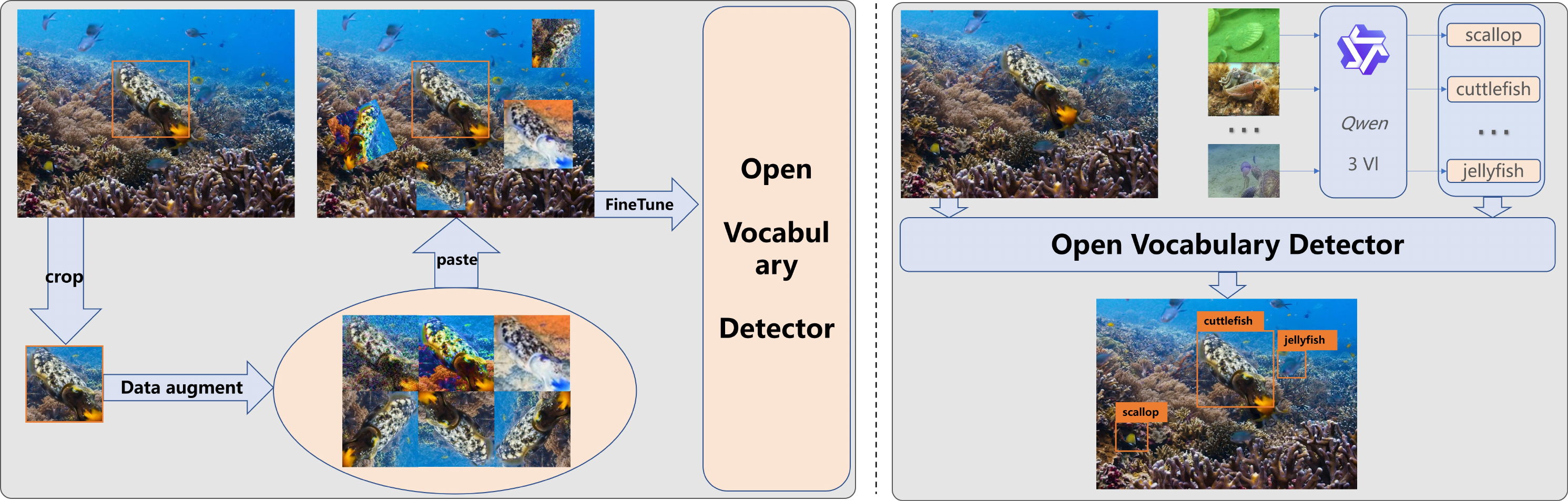}
    }{
        \fbox{\parbox[c][0.22\textheight][c]{\columnwidth}{
        \centering \textbf{Figure Placeholder (Tuning \& Inference)}\\
        Put \texttt{figure1.pdf} (or .png/.jpg) in \texttt{teams/team9_FDUROILab_Lenovo/} and check filename.
        }}
    }
    \caption{Overview of their efficient tuning and inference.}
    \label{fig:tuning}
\end{figure}

Given a k-shot setting, where k represents the number of provided object samples, they employ a structured fine-tuning pipeline, which is shown in Figure~\ref{fig:tuning}. (1) \textbf{Object Cropping and Augmentation.} Using the provided bounding boxes of k-shot examples, they first crop the target objects from the original images. The cropped objects are then subjected to various data augmentation techniques, including flipping, rotation, grayscale conversion, and other transformations, to introduce diversity and improve generalization. (2) \textbf{Object Rescaling and Random Pasting.} They randomly rescale the augmented objects to different sizes and paste these transformed objects onto the original images at different locations. This step simulates new object placements and enhances the model’s robustness to variations in object appearance and context. (3) \textbf{Fine-Tuning with Augmented Data.} They finetune the open-vocabulary detection model with the augmented images. This enables the vision components of the detector to better adapt to objects in the target domain, even with minimal labeled examples. Additionally, the augmented data effectively increases the number of training samples, mitigating the few-shot learning limitation and improving overall detection performance.


    

        

Since their approach utilizes the open-vocabulary detection model, which fundamentally relies on vision-language alignment, it requires access to accurate target category labels during inference, as shown in Figure~\ref{fig:tuning}. To obtain these context-rich labels, they utilize Qwen3-VL~\cite{bai2025qwen3} to generate descriptive textual representations of the target categories. The retrieved target labels from Qwen3-VL are then used as textual input to guide the detection process. Finally, they adopt the detection model to identify and classify objects in the test images based on these enhanced text-based prompts.

\begin{figure}[h]
    \centering
    \def\figPath{teams/team9_FDUROILab_Lenovo/figure2.pdf}
    \IfFileExists{\figPath}{
        \includegraphics[width=\columnwidth]{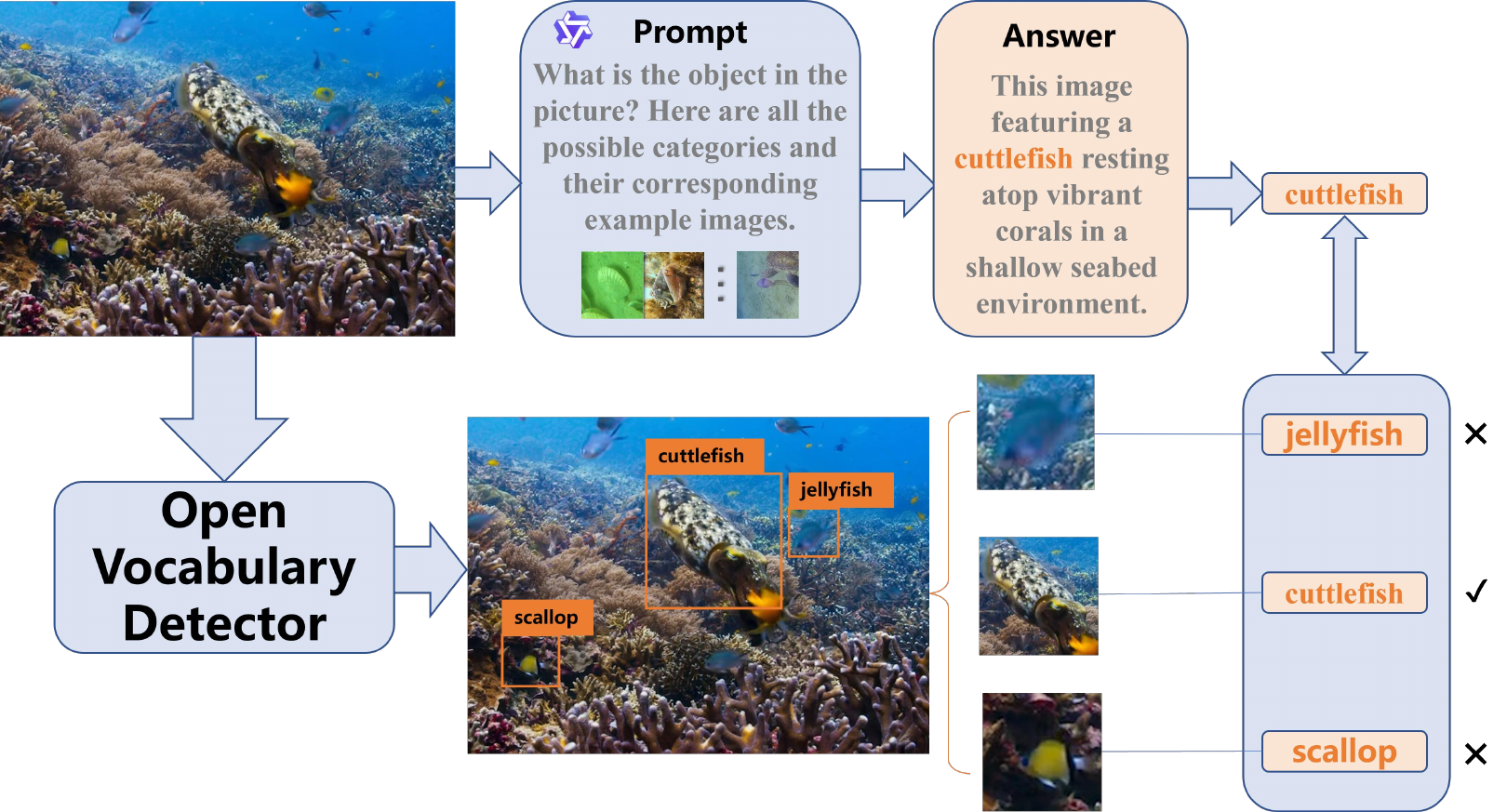}
    }{
        \fbox{\parbox[c][0.22\textheight][c]{\columnwidth}{
        \centering \textbf{Figure Placeholder (Post Processing)}\\
        Put \texttt{figure2.pdf} (or .png/.jpg) in \texttt{teams/team9_FDUROILab_Lenovo/} and check filename.
        }}
    }
    \caption{Post Processing.}
    \label{fig:post}
\end{figure}

Although modern vision-language detectors possess strong generalization capabilities, their performance on the challenging cross-domain test set remains suboptimal in certain cases. Upon further analysis, they found that while the detector can successfully localize most objects, its primary weakness lies in classification errors rather than detection failures. This indicates that the detector still struggles with fine-grained classification when adapting to objects in a new domain. To address this issue, they introduce Qwen3-VL as an auxiliary classifier to refine the final predictions, which is illustrated in Figure~\ref{fig:post}.

Specifically, for each test image, they construct a multimodal prompt comprising the target scene and a set of representative example images for all candidate categories. By leveraging these example images as visual prompts, they instruct Qwen3-VL to describe the objects present in the scene and output a refined list of categories that are likely to appear. After that, they refine the output of the detection model using one of two strategies: (1) \textbf{Filtering.} Remove objects that are classified incorrectly by the detector and are not listed by Qwen3-VL. (2) \textbf{Reclassification.} Assign all detected objects to one of the categories predicted by Qwen3-VL, ensuring consistency between the detected bounding boxes and the high-level scene understanding of the multimodal model. The choice between these two strategies depends on the specific test dataset, as detailed in the Implementation Details. By leveraging Qwen3-VL and visual prompting as a post-processing step, they effectively correct classification errors and enhance the model’s performance on unseen domains, leading to more accurate and reliable object detection results.

\subsubsection{Training Details}

They use the open-vocabulary detection model as the baseline detection model. They utilize Qwen3-VL-235B-A22B~\cite{bai2025qwen3} as their MLLM for label generation and post-processing. Their fine-tuning experiments are conducted on 8 NVIDIA RTX 3090 GPUs, with a batch size of 8 and a base learning rate of 1e-6. During the optimization process, the \texttt{text\_model} is completely frozen, and an \texttt{lr\_mult} of 0.01 is applied to it. They experiment with different numbers of training iterations across datasets and few-shot settings. For dataset1 and dataset3, they fine-tune the model for 30, 50, and 100 batches under 1-shot, 5-shot, and 10-shot settings. For dataset2, no fine-tuning is performed.

To maximize classification accuracy, they adopt dataset-specific post-processing strategies. For dataset1, they reclassify all detected objects into one of the categories predicted by Qwen3-VL. For dataset3, they filter out any detected objects that do not belong to the MLLM-predicted categories. For dataset2, since it contains only a single object category, they do not perform additional classification. However, they apply a specific filtering step: they eliminate overly large detection boxes, which are likely incorrect, as objects in dataset2 are generally small. Besides, they also apply Non-Maximum Suppression (NMS) to suppress redundant or overlapping bounding boxes across all datasets. These dataset-specific strategies ensure that their model achieves optimal performance across different domain shifts and few-shot settings.

\subsection{CDiscover}

\subsubsection{Proposed Method}

\begin{figure}[!htbp]
\centering
\includegraphics[width=1.\linewidth]{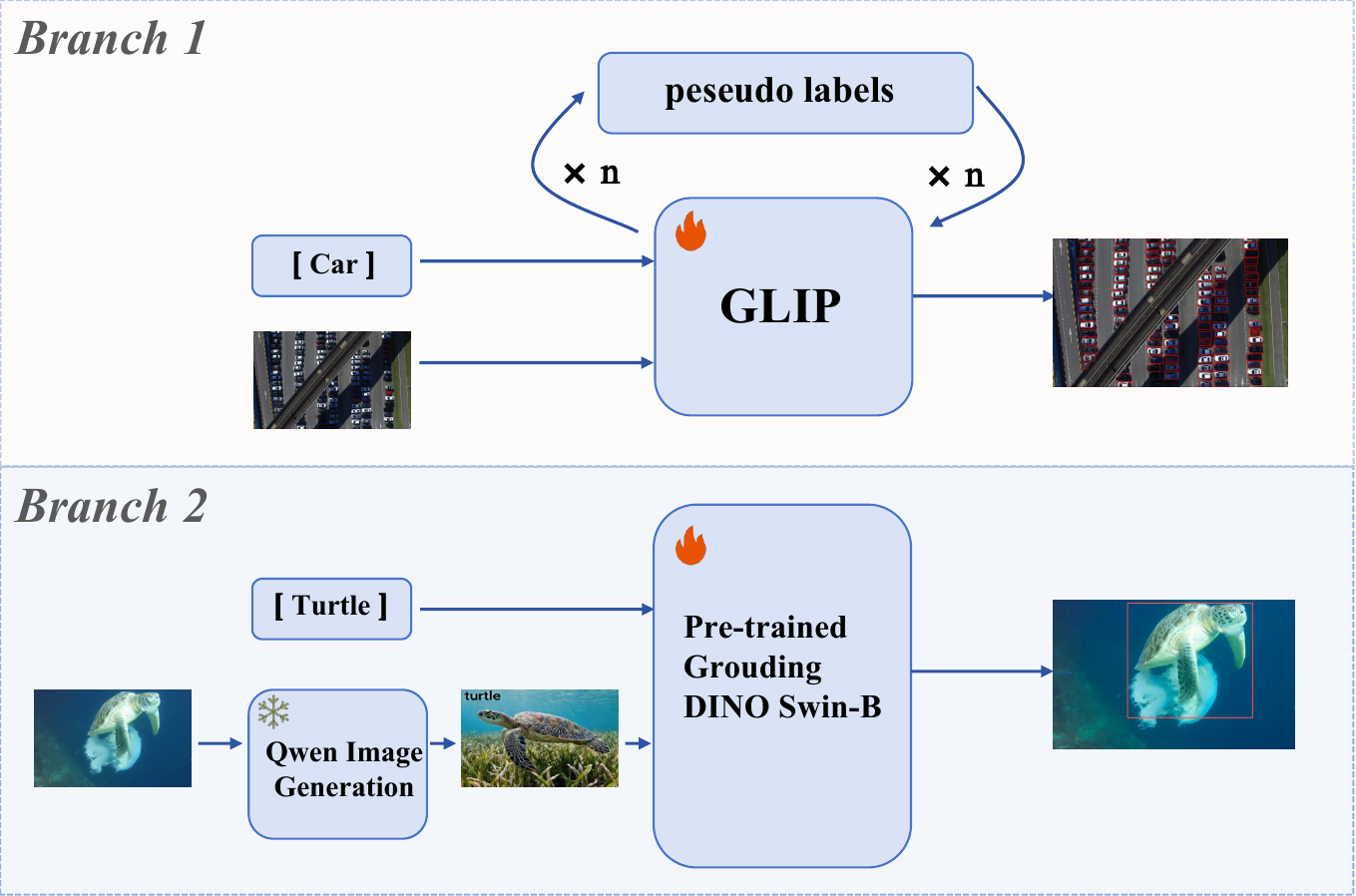} 
\caption{Proposed framework for CDFSOD. 
Branch 1 iteratively refines GLIP with pseudo labels, while Branch 2 uses Qwen image generation to augment training data for Grounding DINO.}
\label{fig:framework5}
\end{figure}

Their method addresses the CDFSOD challenge through a domain-adaptive hybrid strategy, as illustrated in Fig.~\ref{fig:framework5}. For Dataset 1\&3, they leverage the Qwen\cite{wu2025qwen} to synthesize generative data, enriching the feature space for GroundingDINO \cite{liu2024grounding}. For Dataset 2 (a multi-object natural domain with vehicles), they identify that only a single instance is labeled per image despite the presence of multiple targets. To prevent the model from treating unlabeled vehicles as background, they employ a GLIP\cite{li2022grounded}-based pseudo-labeling pipeline to recover missing annotations, followed by iterative self-training to achieve dense and robust object localization.\\

\noindent \textbf{Key contributions:}
\begin{itemize}
    \item \textbf{Contribution 1:} They introduce a generative augmentation pipeline using Qwen to synthesize domain-specific training samples, effectively mitigating the data scarcity issue in CDFSOD.
    \item \textbf{Contribution 2:} They develop an iterative self-training framework based on GLIP and a robust pseudo-label filtering mechanism to enhance model adaptation in complex target distributions.
\end{itemize}

They utilize two vision-language foundation models, selecting the optimal architecture based on the target domain's attributes. In detail, for Dataset 1 and 3, they utilize GroundingDINO, while for Dataset 2, they employ GLIP.
\begin{itemize}
    \item Backbone: The GLIP-L variant is used, which incorporates Swin-L\cite{liu2021swin} as the visual encoder and BERT\cite{devlin2019bert}. For Grounding DINO, the Swin-B\cite{liu2021swin} backbone is used as the visual encoder and BERT from Hugging Face as the text encoder.
    \item Detector: Grounding DINO follows a DETR-style\cite{carion2020end} detection paradigm with a Transformer encoder-decoder architecture. In contrast, GLIP adopts a two-stage detector, consisting of a Region Proposal Network (RPN)\cite{ren2015faster} and ROI heads for region-level prediction.
\end{itemize}

\subsubsection{Training details}
\begin{itemize}
    \item Training data: GLIP is pre-trained on FourODs\cite{krizhevsky2012imagenet,krishna2017visual,kuznetsova2020open}, GoldG\cite{kamath2021mdetr}, CC3M+12M, and SBU\cite{ordonez2011im2text}, while GroundingDINO is pre-trained on a larger and more diverse set of datasets, including COCO\cite{lin2014microsoft}, Objects365\cite{shao2019objects365}, GoldG, Cap4M, OpenImages\cite{kuznetsova2020open}, ODinW-35\cite{li2022elevater}, and RefCOCO\cite{kamath2021mdetr}.
    \item Optimization: For Dataset 2, they fine-tune GLIP-L initialized from the publicly available \texttt{glip\_large\_model.pth} checkpoint. The learning rate is set to $5\times10^{-5}$, the weight decay is $0.05$, and the batch size is 2. They disable AMP and adopt the default GLIP fine-tuning framework with AutoStep scheduling. The step patience is set to 3 and the auto-termination patience is set to 6. The model is trained for 200 epochs, and the random seed is fixed to 10 for reproducibility. Datasets 1 and 3 are trained on a single NVIDIA A6000 GPU, with 50 training epochs and a fixed batch size of 4. The model architecture is built upon the Grounding DINO framework, with core parameters and module configurations as follows: the default number of queries is 900, and the maximum text token length is set to 256; the text encoder adopts a BERT-based structure equipped with BPE tokenization; both the feature enhancer and cross-modality decoder are stacked with 6 layers, and a deformable attention mechanism is introduced in the image cross-attention module to improve feature extraction accuracy. The loss function consists of classification (or contrastive) loss, box L1 loss, and GIoU loss. Following the settings of the original framework, the Hungarian matching weights are set to 2.0 for classification loss, 5.0 for L1 loss, and 2.0 for GIoU loss, while the final total loss weights are correspondingly set to 1.0, 5.0, and 2.0. In addition, They employ a data augmentation strategy that uses a large model to generate samples from the support set for training.  
    \item Augmentations: They augment the support set using a vision-language model (Qwen-image-2.0-pro)\cite{wu2025qwen}. Specifically, given each support image, Qwen is used to generate semantically consistent variations. These synthesized samples effectively enrich the limited support set and improve generalization in few-shot settings.
    \item Hardware / runtime: 2 × Nvidia A6000
\end{itemize}

\noindent \textbf{Inference details.}
Explain inference-time details that affect performance.
\begin{itemize}
    \item Test-time settings: During inference, the detector is applied in a standard single-model setting without test-time augmentation. Since the task is object detection, non-maximum suppression (NMS) is used to remove highly overlapped predictions and retain the most confident bounding boxes. They follow the default inference pipeline of the corresponding detector framework for prediction filtering and box selection.
    \item Ensemble / TTA: None
    \item Any post-processing: None
\end{itemize}

\subsection{NJUST-KMG}

\subsubsection{Proposed Method}
\noindent \textbf{High-level idea.}
ASTER is a hybrid framework built upon FSOD-VFM and ETS. The FSOD-VFM branch operates in a training-free manner: it generates category-agnostic proposals via UPN, refines regions with SAM2, extracts DINOv2 features, matches them against few-shot support prototypes, and suppresses duplicate detections through graph diffusion. The ETS branch provides a trainable counterpart based on GroundingDINO, Swin-B, and BERT, augmented with strong mixed-image strategies. To bridge the two branches, theyconvert high-confidence FSOD-VFM predictions on unlabeled target-domain images into pseudo annotations and continue training ETS for a few additional epochs. Additionally, Domain-RAG is employed to generate compositional target-domain samples, which is particularly effective in extremely low-shot settings.

\noindent \textbf{Key contributions:}
\begin{itemize}[leftmargin=1.5em]
\item \textbf{Strong training-free baseline:} FSOD-VFM provides a robust non-parametric starting point, reaching nearly 60 mAP on dataset2 in their local evaluation.
\item \textbf{Teacher-to-student adaptation:} High-confidence FSOD-VFM detections are converted into COCO-format pseudo annotations and incorporated into ETS training as supplementary supervision.
\item \textbf{Low-shot data diversification:} Domain-RAG augments appearance diversity, which is especially critical for 1-shot cases where target-domain coverage is most limited.
\end{itemize}
\noindent \textbf{Model Architecture}
\begin{itemize}[leftmargin=1.5em]
\item \textbf{FSOD-VFM branch:} UPN proposals + SAM2 masks + DINOv2 ViT-L/14 features + prototype matching + graph-diffusion score reweighting. For certain dataset3 settings, prototype refinement from high-confidence query predictions is additionally enabled.
\item \textbf{ETS branch:} GroundingDINO with a Swin-B backbone, BERT-base-uncased language encoder, 900 queries, the standard GroundingDINO encoder-decoder architecture, and class text prompts derived from target dataset categories.
\item \textbf{Hybrid bridge:} FSOD-VFM outputs are filtered by dataset-specific score thresholds to produce pseudo annotations, which are then concatenated with the real few-shot annotations for an additional stage of ETS fine-tuning.
\end{itemize}

\subsubsection{Training Details}
\begin{itemize}[leftmargin=1.5em]
\item \textbf{Training data:} Official challenge few-shot splits of dataset1, dataset2, and dataset3; pseudo annotations generated from FSOD-VFM predictions on unlabeled target-domain images; Domain-RAG synthetic samples merged into \texttt{*\_shot\_rag\_aug.json}.
\item \textbf{Few-shot setting:} 1-shot, 5-shot, and 10-shot. During pseudo-label continuation, real few-shot annotations are upsampled to prevent them from being overwhelmed by pseudo annotations. The repeat factors are 32 for dataset1, 64 for dataset2, and 32 for dataset3.
\item \textbf{Optimization:} The ETS baseline uses AdamW with a learning rate of \(1\times10^{-4}\), weight decay of \(1\times10^{-4}\), batch size 2, and gradient clipping at 0.1. Training schedules are dataset-specific: 30 epochs for dataset1 1/5-shot, 10 epochs for dataset1 10-shot, 7 epochs for dataset2, 25 epochs for dataset3 1-shot, and 20 epochs for dataset3 5/10-shot.
\item \textbf{Pseudo-label continuation:} ETS is initialized from the best baseline checkpoint and further fine-tuned for 4 epochs with a reduced learning rate of \(2\times10^{-5}\).
\item \textbf{Augmentations:} CachedMosaic, CachedMixUp, HSV jittering, random flip, multi-scale resize, and random crop are applied in the ETS training. Domain-RAG contributes additional compositional target-domain samples.
\item \textbf{Hardware / runtime:} All experiments are conducted on a single NVIDIA RTX 4090 GPU. FSOD-VFM is training-free, with computation primarily spent on proposal generation, mask extraction, and feature caching. ETS baseline training and pseudo-label fine-tuning share the same single-GPU setup.
\end{itemize}

\noindent \textbf{Inference Details}
\begin{itemize}[leftmargin=1.5em]
\item \textbf{FSOD-VFM:} DINOv2 ViT-L/14, SAM2 Large, UPN Large, 15 graph-diffusion steps, \(\alpha=0.3\), \(\lambda=0.5\). They retain 100 proposals by default, increasing to 200 for selected dataset2 and dataset3 settings, with a minimum proposal threshold of 0.01 or 0.005 depending on the dataset.
\item \textbf{Score fusion:} Class scores are computed from support prototypes, optionally fused with proposal confidence, and then reweighted via graph diffusion. For dataset3 1-shot and 10-shot, prototype refinement from high-confidence query predictions is additionally applied.
\item \textbf{ETS:} Inference uses the best checkpoint selected by local mAP. The detector retains up to 200 or 300 predictions per image depending on the dataset configuration.
\item \textbf{Final selection:} For each dataset-shot setting, they submit the best-performing branch among ETS baseline, ETS with FSOD-VFM pseudo labels, and ETS with Domain-RAG, determined by their local model-selection protocol.
\end{itemize}

\subsection{Earth-insights}

\subsubsection{Proposed Method}
\begin{figure}
    \centering
    \def\figPath{teams/team18_earth-insights/images/earth_insights.pdf}
    \IfFileExists{\figPath}{
        \includegraphics[width=\linewidth]{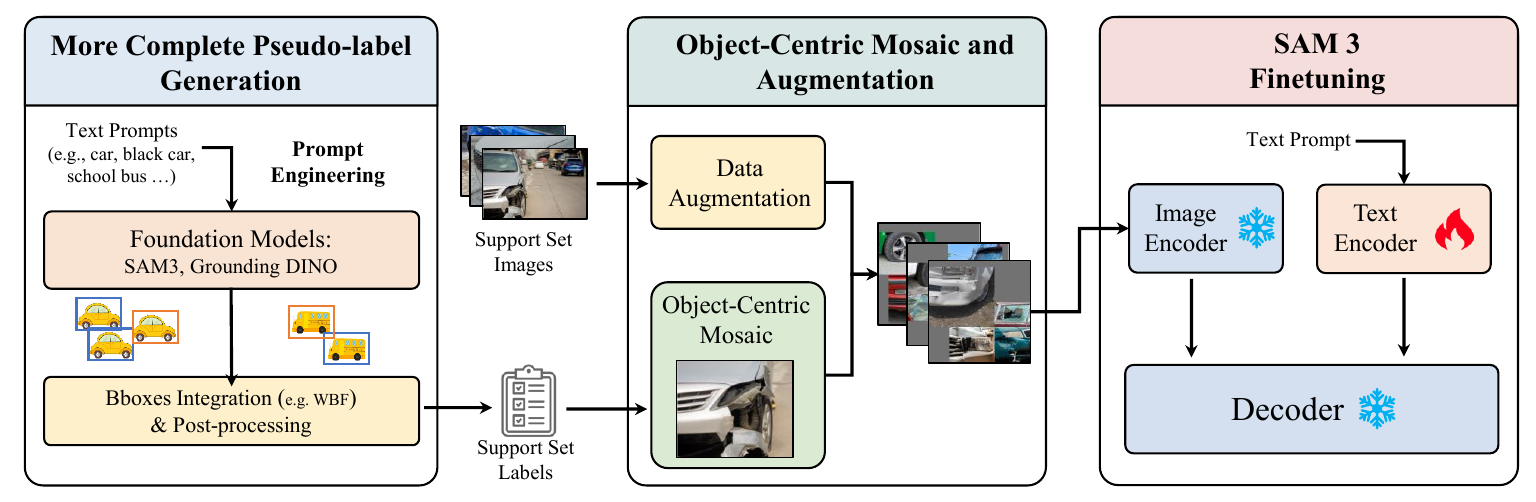}
    }{
        \fbox{\parbox[c][0.22\textheight][c]{\linewidth}{
        \centering \textbf{Figure Placeholder (earth-insights)}\\
        Put \texttt{earth_insights.pdf} (or .png/.jpg) in \texttt{teams/team18_earth-insights/images/} and check filename.
        }}
    }
    \caption{Team earth-insights: overview of the training process.}
    \label{fig:earth_insights}
\end{figure}

To tackle the CD-FSOD task under the open-source setting, this team proposes a data-centric fine-tuning pipeline (Fig.~\ref{fig:earth_insights}) built upon powerful foundation models, specifically SAM3~\cite{carion2025sam} and Grounding DINO~\cite{liu2024grounding}. Recognizing that the quality and diversity of the support set are bottlenecks in few-shot scenarios, the proposed method focuses on more complete pseudo-label generation, targeted data synthesis, and dataset-specific adaptation strategies. 
The overall pipeline consists of the following key steps:

\noindent
\textbf{High-Quality Support Label Generation.} To maximize the utility of the limited support set, this team leverages the zero-shot capabilities of SAM3 and Grounding DINO. Some prompt engineering is applied to extract accurate initial bounding boxes. For instance, in the parking lot dataset (Carpk), specific textual prompts such as ``school bus'' or ``black car'' are utilized to guide the foundation models effectively. To refine these initial predictions, the Weighted Boxes Fusion (WBF)~\cite{solovyev2021weighted} algorithm, along with carefully designed post-processing techniques, is employed to ensemble the results, yielding more complete support labels for subsequent training.

\noindent
\textbf{Object-Centric Mosaic and Augmentation.} Since the generated labels may still contain imperfections and the support samples are extremely scarce, this team introduces an ``Object-Centric Mosaic'' data synthesis strategy. Instead of randomly cropping images for the standard Mosaic augmentation~\cite{bochkovskiy2020yolov4}, this approach specifically crops regions surrounding the annotated targets. This ensures that the synthesized samples maintain high information density and context relevance. Additional standard data augmentations are also incorporated to further expand sample diversity and alleviate overfitting.

\noindent
\textbf{Dataset-Specific Strategy and Validation.} The team observed a high variance in training results due to sample scarcity. Following~\cite{pan2025enhance}, an optimized target domain validation set is constructed to monitor the training process and select the best checkpoints. Furthermore, empirical results revealed that the native SAM3 exhibits significantly different adaptation capabilities across various datasets. For example, on the Carpk, which predominantly contains small cars, SAM3 achieves a zero-shot mAP of over 50\%. Conversely, on the CarDD, which features objects with ambiguous boundaries (\textit{e.g.}, ``scratch''), the zero-shot mAP drops below 10\%. Consequently, the specific data processing and augmentation strategies are dynamically adjusted based on the object attribute and zero-shot performance characteristics of each specific dataset.

\subsubsection{Training Details}

The proposed method uses SAM3 as the primary foundation model for the fine-tuning stage. To maintain training efficiency within a limited timeframe, the team utilized the default SAM3 fine-tuning script, wherein only the parameters of the text encoder are updated while other parts remain frozen. This team encourages future participants to explore more comprehensive parameter-efficient fine-tuning schemes for SAM3. All experiments are conducted on a single NVIDIA A100 GPU. During the training phase, strict adherence to the few-shot protocol is maintained; no external data is used other than the provided support set images and the pre-trained weights of the foundation models.

\noindent
\textbf{Additional Explorations:} For the underwater dataset scenarios, this team also experimented with an underwater-specific open-vocabulary segmentation model~\cite{li2025maris} to bridge the distinct domain gap. Although these results were not included in the final submission due to time constraints, further exploration is valuable in future research.

\subsection{Intellindust AI Lab}
\subsubsection{Proposed Method}

\begin{figure}[!htbp]
\centering
\includegraphics[width=1.\linewidth]{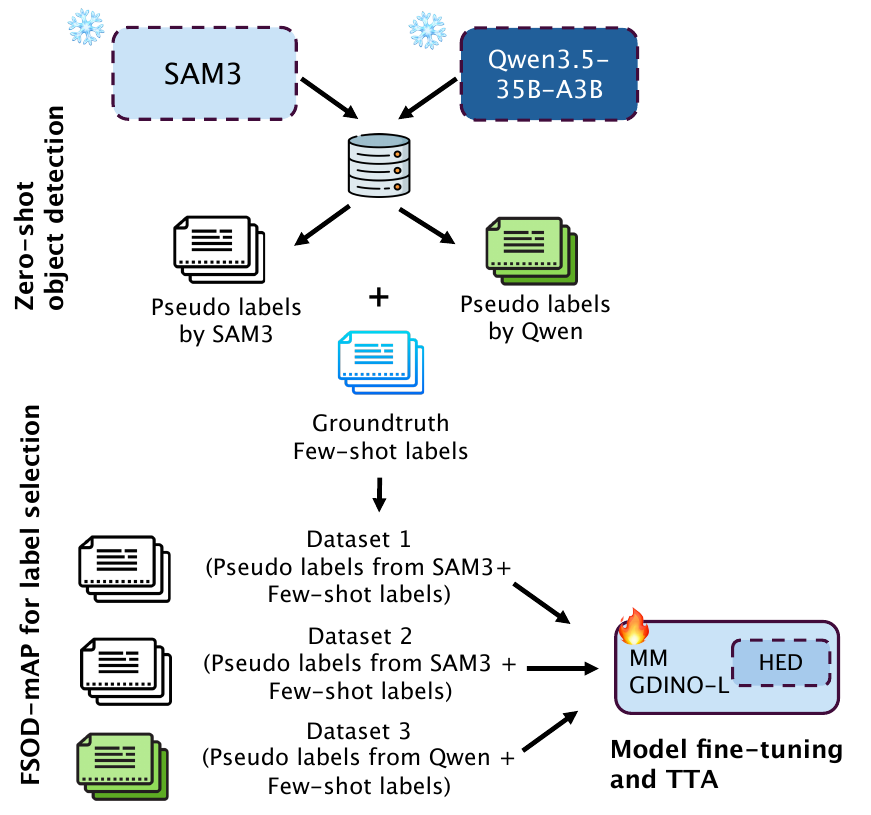}
\caption{Overall pipeline of \texttt{ZEP}.}
\label{fig:method6}
\end{figure}

They generate pseudo-labels for three datasets in a zero-shot manner using Qwen3.5-35B-A3B~\cite{qwen3.5} and SAM3~\cite{carion2026sam}, where only class-name text prompts are provided as input. The overall framework is given in Fig.~\ref{fig:method6}.

To automatically select more reliable pseudo-labels for each dataset, they propose a metric termed FSOD-mAP. Specifically, for each dataset, they compute the IoU between pseudo-labels and the few-shot ground-truth boxes (1/5/10-shot). Predictions with IoU $\leq$ 0.3 and matching class labels are filtered out to suppress noisy false positives. They then compute mAP on the remaining predictions, which serves as a proxy for pseudo-label quality. Based on this criterion, they select the pseudo-label source with higher FSOD-mAP (Dataset1 and Dataset2 favor SAM3, while Dataset3 favors Qwen3.5). Additionally, for SAM3-generated pseudo-labels, they further discard predictions with confidence scores lower than 0.8 (Qwen3.5 does not provide confidence scores).

Next, they merge pseudo-labels with the few-shot ground-truth annotations. To avoid redundancy, pseudo-labels that have an IoU greater than 0.8 with ground-truth boxes of the same class are removed.

They explore two strategies for constructing training and validation sets: \textit{Strategy 1}: The merged dataset (pseudo-labels + few-shot annotations) is split into training and validation sets with a ratio of 8:2.
\textit{Strategy 2}: The merged training set is used for training, while pseudo-labels generated on the original test set (without fusion, due to the absence of ground truth) are used as the validation set.

They then fine-tune the two MMGroundingDINO-L~\cite{zhao2024open} models with the Hybrid Ensemble Decoder (HED)~\cite{yu2026acloser}, which is proposed by their team (accepted by CVPR 2026, forthcoming). During inference, they apply test-time augmentation (TTA) with horizontal flipping and Soft-NMS for each model. Finally, predictions from the two models (trained under the two strategies) are combined using Soft-NMS to produce the final results. 
\subsection{SAIDA}

\subsubsection{Proposed Method}

\noindent\textbf{High-level Idea. }Cross-Domain Few-Shot Object Detection (CD-FSOD) \cite{fu2024cross} presents a formidable challenge in computer vision, requiring models to adapt to novel target distributions with extremely sparse supervision—typically limited to 1, 5, or 10-shot annotations. A critical vulnerability in standard training pipelines for CD-FSOD is the inherent risk of false negatives; when training on such sparse labels without preliminary refinement, the model often incorrectly learns to categorize unannotated objects as background, leading to catastrophic interference during domain transfer.

\begin{figure}[h]
\centering
\def\figPath{teams/team5_SAIDA/method_figure.png}
\IfFileExists{\figPath}{
  \includegraphics[width=\columnwidth]{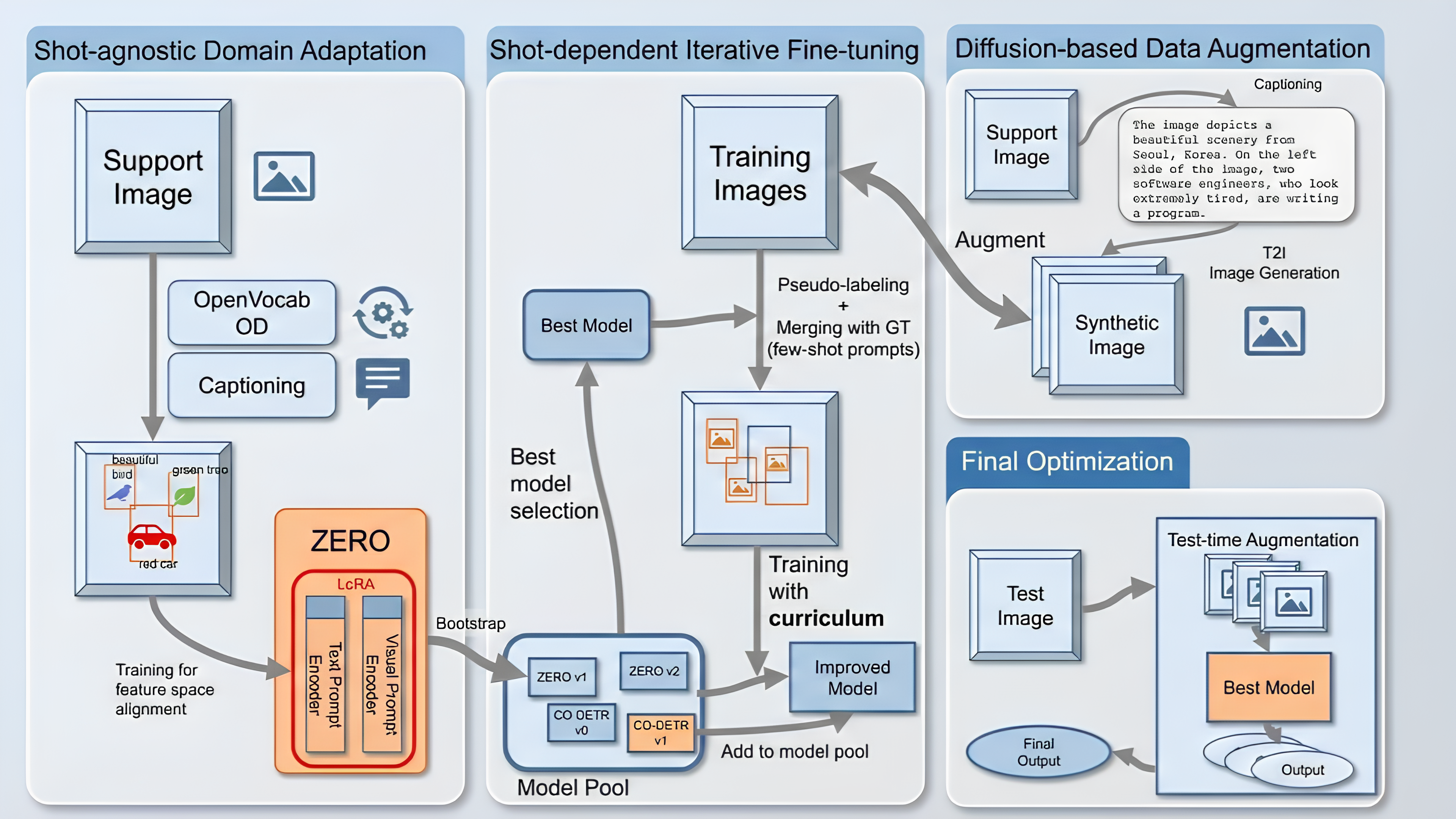}
}{
  \fbox{\parbox[c][0.22\textheight][c]{\columnwidth}{
  \centering \textbf{Method Figure Placeholder}\\
  Put \texttt{method\_figure.png} (or .pdf/.jpg) in \texttt{teams2026/team5_SAIDA/} and check filename.
  }}
}
\caption{Overall pipeline of \texttt{SAIDA}}
\label{fig:method8}
\end{figure}

Synthetic-Augmented Iterative Domain Adaptation (SAIDA) addresses these constraints through a four-phase pipeline, as demonstrated in Fig.~\ref{fig:method8}. By progressing from label-agnostic adaptation to iterative pseudo-labeling and generative data expansion, SAIDA optimizes the latent space for domain-specific distributions and maximizes model precision in data-scarce environments.

\subsubsection{Module Details}


\noindent\textbf{Shot-agnostic domain adaptation.}
The initial phase focuses on adapting the model to the target domain's visual distribution without relying on the provided challenge labels. This ensures the architecture captures the underlying semantics of the new domain before category-specific fine-tuning begins.

\begin{itemize}
    \item Foundation model selection: They utilize ZERO\cite{spbaizero}, a vision foundation model developed by Superb AI. ZERO is a zero-shot/few-shot object detection model specifically engineered for visual grounding tasks and supports both text and visual prompts, making it an ideal candidate for zero-shot domain initialization.
    \item Noun phrase extraction Pipeline: To identify potential objects in the target domain, they employ an ensemble of open-vocabulary detectors—including Grounding DINO\cite{groundingdino}, YOLO-E\cite{yoloe}, and SAM3\cite{sam3}. These models use target category names as text prompts to distinguish objects from the background. The resulting detections are processed by  Qwen3-VL \cite{qwen3vl} to extract descriptive semantic noun phrases, which provide richer context than primitive category labels.
    \item Alignment and reconstruction: The model is optimized using (bounding box, noun phrase) pairs. They employ a dual-loss objective that incorporates both CLIP\cite{clip} and CapPa\cite{cappa} losses without modifying the model architecture of ZERO. This approach ensures global semantic alignment in the embedding space while simultaneously performing conditional reconstruction to stabilize the vision-language projection.
    \item Efficiency via parameter-efficient fine-Tuning: To maintain the predictive performance of the pretrained weights while adapting to new domain semantics, they apply Low-Rank Adaptation (LoRA)\cite{lora} specifically to the prompt encoders. This preserves the generalization capabilities of the foundation model while minimizing the computational footprint of the adaptation.
\end{itemize}

\noindent\textbf{Iterative shot-dependent fine-tuning.}
Following the shot-agnostic adaptation, the methodology transitions to leveraging the provided challenge annotations with a shot-dependent fine-tuning through a iterative process.
\begin{itemize}
    \item Prompt engineering: They utilize human evaluation and feedback from evaluation servers to identify the optimal prompt types for each (dataset, shot) pair. This ensures that the initial pseudo-labels are generated using the most effective semantic cues for the specific domain.
    \item Mathematical threshold optimization: They calculate class-wise score thresholds to optimize the pseudo-labeling process by F-score optimization. By weighing precision higher than recall, they mitigate semantic drift and prevent the propagation of label noise into the training set. As model reliability improves over iterations, they progressively shift the balance from precision to recall, since the risk of introducing noisy pseudo-labels decreases while the benefit of covering more object instances increases.
    \item Label merging strategy: They employ a class-agnostic Non-Maximum Suppression (NMS) to integrate pseudo-labels with Ground-Truth (GT) annotations. In this logic, GT labels are assigned a confidence score of 1.0, ensuring they take priority during the merging process.
    \item Model evolution: This phase facilitates the continual learning of both the prompt encoder finetuned ZERO and the Object365\cite{object365} pretrained Co-DETR\cite{codetr}, with each round refining the quality of the training signals. They follow the default training configurations (e.g., data augmentation, optimizer) provided in the Transformers\cite{transformers} and MMDetection\cite{mmdetection} libraries for ZERO and Co-DETR, respectively\footnote{\href{https://github.com/huggingface/transformers/blob/main/examples/pytorch/object-detection/run_object_detection_no_trainer.py}{ZERO training config}, \href{https://github.com/open-mmlab/mmdetection/blob/main/projects/CO-DETR/configs/codino/co_dino_5scale_swin_l_16xb1_16e_o365tococo.py}{Co-DETR training config}}. For the detailed model architectures, please refer to their technical reports.
\end{itemize}

\noindent\textbf{Diffusion-based data augmentation.}
To circumvent the performance saturation often observed with limited datasets, they expand the training data by a factor of 10 using an image generative model.

\begin{itemize}
    \item Captioning and T2I fine-tuning: Qwen3-VL \cite{qwen3vl} is used to generate high-fidelity captions for the target domain images. These pairs are used to fine-tune Qwen-Image\cite{qwenimage} (via the DiffSynth-Studio library\cite{diffsynth2024}) using LoRA\cite{lora}.
    \item Synthetic scaling: They generate synthetic images using the fine-tuned Qwen-Image\cite{qwenimage} and annotate them with pseudo-labels produced by ZERO\cite{spbaizero} or Co-DETR\cite{codetr} from the previous round. These synthetic samples are incrementally incorporated into subsequent training rounds, improving the model’s robustness to domain shifts and enhancing overall generalization.
\end{itemize}

\noindent\textbf{Final optimization.}
In the final phase, they conduct empirical tuning to maximize discriminative performance.
\begin{itemize}
    \item Curriculum data augmentation: Rather than relying on generative augmentations, they tried image processing-based augmentation techniques. In particular, they employ color augmentations, which is commonly used as strong augmentations in semi-supervised learning, to improve robustness during training \cite{guo2022scale}.
    \item Test-time augmentation: During inference, they employ multi-scale resizing (×0.75, ×1.0, ×1.25) combined with horizontal flipping.
\end{itemize}

They propose a parameter-efficient domain adaptation approach that preserves pretrained knowledge while mitigating catastrophic forgetting. In addition, they develop an iterative self-improvement procedure that operates without direct human-provided labels, ensuring full compliance with the challenge rules by minimizing human involvement that could be considered unfair or misleading in FSOD settings. Furthermore, beyond exploring image processing-based augmentation policies, they demonstrate that incorporating synthetic data generated by recent diffusion models provides substantial benefits in CD-FSOD scenarios.

\subsection{KLETech-CEVI}

\subsubsection{Proposed Method}

\begin{figure}[t]
\centering
\includegraphics[width=1.\linewidth]{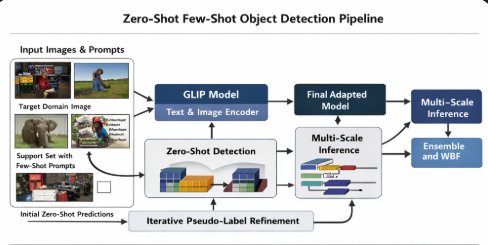}
\caption{Overview of the proposed pipeline.}
\label{fig:cevi}
\end{figure}

The KLETech-CEVI team addresses Cross-Domain Few-Shot Object Detection (CD-FSOD) using a pseudo-label driven vision-language grounding framework built on GLIP. The approach leverages the strong semantic priors of vision-language models while mitigating their localization limitations under domain shift. As summmarized in Fig.~\ref{fig:cevi}, the overall pipeline consists of three stages:
\begin{enumerate}
    \item Zero-shot detection using GLIP
    \item Pseudo-label generation
    \item Iterative model adaptation via fine-tuning
\end{enumerate}

Given initial predictions $B$, high-confidence pseudo-labels are selected as:
\begin{equation}
\hat{B} = \{ b_i \in B \mid s_i > \tau \}
\end{equation}

The KLETech-CEVI team applies multiple filtering strategies to improve pseudo-label quality, including confidence thresholding, non-maximum suppression (NMS), and class-consistency filtering.

The model is then iteratively refined using the following objective:
\begin{equation}
\mathcal{L} = \mathcal{L}_{cls} + \lambda \mathcal{L}_{loc}
\end{equation}

\begin{equation}
\mathcal{L}_{loc} = \mathcal{L}_{SmoothL1} + \beta \mathcal{L}_{GIoU}
\end{equation}

This design improves both classification robustness and localization accuracy.

To further enhance performance, multi-scale test-time augmentation is applied at resolutions $\{640, 800, 1000\}$, and predictions are fused to improve robustness to object scale variation.

\begin{algorithm}[t]
\caption{Pseudo-Label Driven Adaptation}
\begin{algorithmic}[1]
\STATE Initialize model $M$
\FOR{each iteration}
\STATE Generate predictions $B$
\STATE Filter pseudo-labels $\hat{B}$
\STATE Fine-tune model
\ENDFOR
\end{algorithmic}
\end{algorithm}

\begin{figure}[t]
\centering
\includegraphics[width=1.\linewidth]{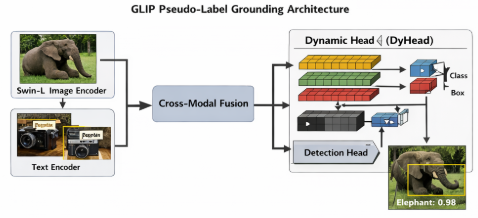}
\caption{GLIP-based architecture.}
\label{fig:glip-framework}
\end{figure}

As detailed in Fig.~\ref{fig:glip-framework}, the architecture is based on GLIP and includes:
\begin{itemize}
    \item Swin-L backbone~\cite{liu2021swin}
    \item GLIP text encoder~\cite{li2022glip}
    \item DyHead detection head
\end{itemize}

The KLETech-CEVI team demonstrates that combining zero-shot grounding with iterative pseudo-label refinement effectively bridges domain gaps and significantly improves detection performance, especially under low-shot settings.

\subsubsection{Training Details}

The KLETech-CEVI team trains the model using large-scale vision-language datasets, including Conceptual Captions, SBU Captions, Visual Genome~\cite{krishna2017vg}, MS COCO~\cite{lin2014coco}, and Objects365.
For optimization, the following loss functions are employed:
\begin{itemize}
    \item Focal Loss~\cite{lin2017focal} for handling class imbalance and improving recall
    \item Generalized IoU (GIoU) Loss~\cite{rezatofighi2019giou} for enhanced localization accuracy
\end{itemize}

The iterative pseudo-labeling process plays a central role in training. High-confidence predictions are progressively incorporated as supervision, enabling domain adaptation without requiring additional manual annotations.
Experimental results show consistent improvements across multiple datasets and shot settings. In particular, the method achieves more than 2$\times$ improvement in the 1-shot scenario, demonstrating strong effectiveness in low-data regimes.
Ablation studies conducted by the KLETech-CEVI team indicate that:
\begin{itemize}
    \item Pseudo-labeling provides the largest performance gain
    \item Focal Loss improves detection of hard examples
    \item GIoU Loss enhances bounding box quality
    \item Multi-scale inference improves robustness to object size variation
\end{itemize}

Despite strong performance, the approach depends on pseudo-label quality and introduces additional computational overhead due to iterative fine-tuning. Performance may also degrade for rare categories with weak semantic alignment, and the confidence threshold $\tau$ requires careful tuning.
Overall, the training strategy effectively leverages vision-language priors and pseudo-label driven adaptation to achieve robust cross-domain generalization.

\subsection{Manifold}
\subsubsection{Proposed Method}

\begin{figure}[!htbp]
\centering
\def\figPath{teams/team15_Manifold/fig1.png}
\IfFileExists{\figPath}{
  \includegraphics[width=1.\linewidth]{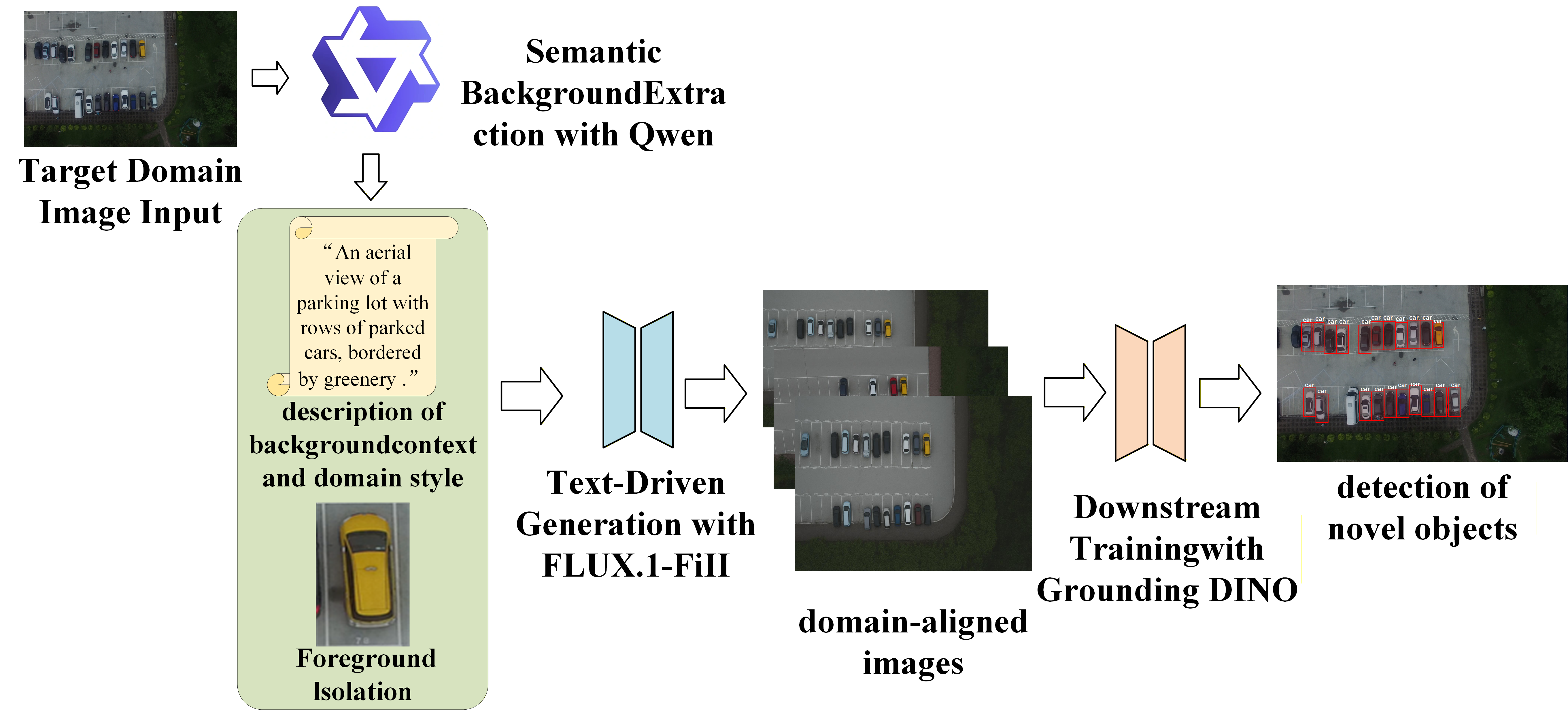}
}{
  \fbox{\parbox[c][0.22\textheight][c]{0.95\linewidth}{
  \centering \textbf{Method Figure Placeholder}\\
  Put \texttt{fig1.png} (or .pdf/.jpg) in \texttt{teams/team15_Manifold/} and check filename.
  }}
}
\caption{Overall pipeline of \texttt{Multimodal Prompt-Driven Diffusion Augmentation (MPDA)}. }
\label{fig:method15}
\end{figure}

\noindent \textbf{Key contributions:}
\begin{itemize}
  \item \textbf{Contribution 1:} They design a targeted data augmentation pipeline that utilizes Qwen3-VL for background prompt extraction and Flux.1 fill for background inpainting, significantly expanding the diversity of few-shot training samples while preserving accurate foreground bounding boxes.
  \item \textbf{Contribution 2:} They provide a strictly source-free alternative to retrieval-based methods like Domain-RAG; by relying entirely on generative synthesis rather than source-domain retrieval, their method eliminates the dependency on source datasets and avoids source-domain bias.
  \item \textbf{Contribution 3:} They successfully fine-tune Grounding DINO (Swin-B backbone) on the synthesized dataset, demonstrating that integrating diffusion-based background inpainting with vision-language detectors practically bridges the domain gap in cross-domain few-shot settings.
  \end{itemize}

\noindent \textbf{Model architecture}
\begin{itemize}
    \item Backbone: Swin-B.
    \item Detector: Grounding DINO.
    \item Additional components: Qwen3-VL  and Flux.1 Fill , integrated as off-line generative data augmentation modules prior to the detection pipeline.
    \end{itemize}

\subsubsection{Training details}
\begin{itemize}
    \item Training data: The training is comprehensively conducted across three distinct target-domain datasets. For each dataset, they synthesize augmented training manifolds based entirely on the provided few-shot support sets using their generative background inpainting pipeline.
    \item Few-shot setting: They systematically evaluate their method under 1-shot, 5-shot, and 10-shot settings for all three datasets, culminating in 9 independent experimental configurations. The support selection strictly follows the official challenge splits.
    \item Optimization: Consistent across all 9 configurations, they use the AdamW optimizer (Initial LR = $1 \times 10^{-4}$, weight decay = $1 \times 10^{-4}$) with $L_2$ gradient clipping (max norm = 0.1). A layer-wise learning rate decay is applied (Swin-B backbone LR multiplier = 0.1)
    \item Augmentations: A robust spatial and color augmentation pipeline is applied, including CachedMosaic (prob=0.6, 640x640), YOLOXHSVRandomAug, RandomFlip (prob=0.5), CachedMixUp (prob=0.3), and Large-Scale Jittering (RandomChoice multi-scale resize between 480x1333 and 800x1333, coupled with absolute range RandomCrop).
    \item Hardware / runtime: All experiments were executed on a single NVIDIA RTX 4090 GPU (24GB VRAM). The training time for each of the 9 configurations is approximately 8 hours.
    \end{itemize}

\noindent \textbf{Inference details}
\begin{itemize}
    \item Test-time settings: {Grounding DINO default settings. Since test set annotations were unavailable for local validation, they saved multiple model checkpoints during the fine-tuning phase and generated inference results for each. Their final submission was determined by selecting the outputs that achieved the highest scores on the official evaluation server.}
    \item Ensemble / TTA: \texttt{None.}
    \item Any post-processing: \texttt{None.}
    \end{itemize}

\subsection{QiFans}
\subsubsection{Proposed Method}

\begin{figure}[!htbp]
\centering
\def\figPath{teams/team21_QiFans/images/pipeline.pdf}
\IfFileExists{\figPath}{
  \includegraphics[width=0.95\linewidth]{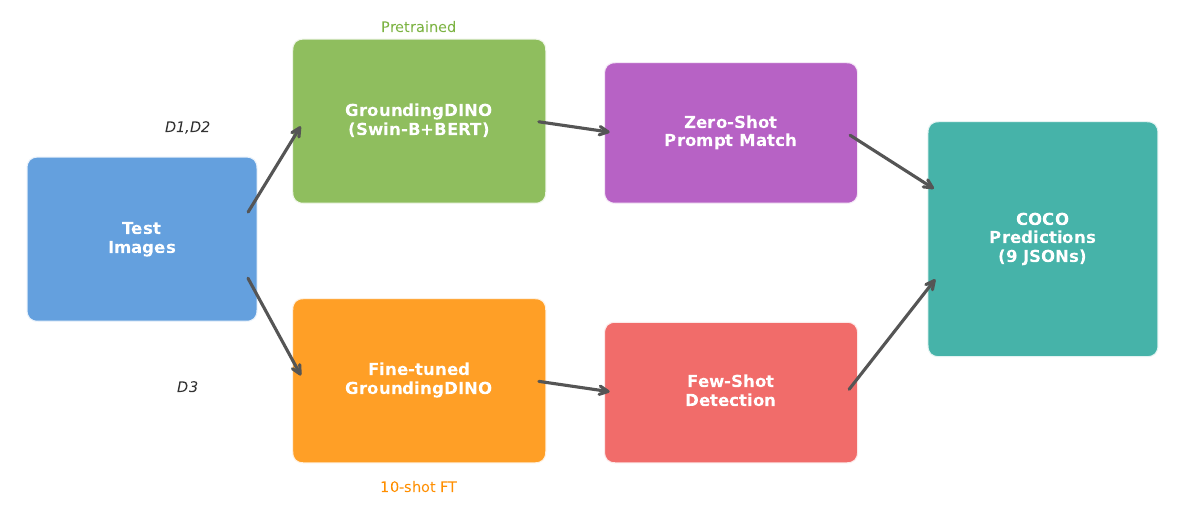}
}{
  \fbox{\parbox[c][0.22\textheight][c]{0.95\linewidth}{
  \centering \textbf{Method Figure Placeholder}\\
  Put \texttt{pipeline.pdf} (or .png/.jpg) in \texttt{teams/team21_QiFans/images/} and check filename.
  }}
}
\caption{Overall pipeline of GDino-FT. For D1 (underwater) and D2 (car), they use pretrained GroundingDINO with optimized text prompts (zero-shot). For D3 (car damage), they fine-tune the model on the 10-shot support set while freezing the BERT text encoder.}
\label{fig:method9}
\end{figure}

\noindent \textbf{High-level idea:} They use GroundingDINO (Swin-B) as a unified foundation model for all three target domains. Their key insight is that \emph{different target domains require fundamentally different strategies}: zero-shot prompt engineering for domains containing common objects (D1: underwater, D2: car), and few-shot fine-tuning for specialized fine-grained domains where text-image alignment is weak (D3: car damage). For D1, they craft descriptive multi-synonym prompts to improve recall on uncommon marine categories. For D3, they fine-tune the visual backbone and detection head on the 10-shot support set with heavy data augmentation while freezing the BERT text encoder. Model architecture is given in Fig.~\ref{fig:method9}.

\noindent \textbf{Key contributions:}
\begin{itemize}
    \item \textbf{Domain-adaptive strategy selection:} Automatically choosing between zero-shot and fine-tuning based on domain characteristics, achieving strong performance across diverse target domains.
    \item \textbf{Optimized prompt engineering:} Descriptive multi-synonym prompts with phrase-to-class mapping doubled D1 mAP (11.23 $\to$ 23.42).
    \item \textbf{Effective few-shot fine-tuning:} Fine-tuning GroundingDINO on just 60 images improved D3 from 9.12 to 36.08 mAP (+26.96), with heavy augmentation preventing overfitting.
\end{itemize}

\noindent \textbf{Model architecture}
\begin{itemize}
    \item Backbone: Swin-B (pretrained on Objects365 + GoldG + Cap4M via GroundingDINO)
    \item Detector: GroundingDINO --- DINO-based open-set detector with cross-modality text-image fusion
    \item Text encoder: BERT-base-uncased (frozen during fine-tuning)
    \item Total parameters: $\sim$172M ($\sim$62M trainable when BERT is frozen)
    \item Foundation models used: GroundingDINO, BERT-base-uncased (2 total)
\end{itemize}

\subsubsection{Training details}
\begin{itemize}
    \item Training data: Only the officially provided 10-shot support set for D3 (60 images, 60 annotations, 6 categories). D1 and D2 use zero-shot only (no support set training).
    \item Few-shot setting: 10-shot for D3. Support images used as-is from the provided JSON annotations.
    \item Optimization: AdamW optimizer, backbone LR=$1\times10^{-5}$, head LR=$5\times10^{-5}$, weight decay=0.05, cosine schedule with 100-step warmup, 200 epochs, batch size 2, mixed precision (FP16), gradient clipping (max norm 0.1)
    \item Loss: Hungarian matching with focal loss ($\alpha$=0.25, $\gamma$=2.0) + L1 loss + GIoU loss, weighted 2:5:2
    \item Augmentations: Random horizontal flip, multi-scale resize (480--800, max 1333), random crop (384--600), color jitter ($p$=0.8), Gaussian blur ($p$=0.3)
    \item Hardware / runtime: 1$\times$ NVIDIA H800 80GB GPU, D3 fine-tuning $\sim$30 minutes
\end{itemize}

\noindent \textbf{Inference details}
\begin{itemize}
    \item D1 (Underwater): Zero-shot, optimized multi-synonym prompt, box\_threshold=0.10, text\_threshold=0.10. Phrase-to-class mapping (e.g., ``sea cucumber'' $\to$ ``holothurian'').
    \item D2 (Car): Zero-shot, simple prompt ``car .'', box\_threshold=0.15, text\_threshold=0.15.
    \item D3 (Car damage): Fine-tuned model, box\_threshold=0.10, text\_threshold=0.10.
    \item Ensemble / TTA: None
    \item Post-processing: Phrase-to-class mapping only. No NMS beyond GroundingDINO's built-in processing.
\end{itemize}




\noindent \textbf{Ablation and analysis}

\begin{table}[t]\scriptsize
\centering
\begin{tabular}{lcccc}
\toprule
\textbf{Configuration} & \textbf{D1} & \textbf{D2} & \textbf{D3} & \textbf{Total} \\
\midrule
(a) ZS, basic prompts & 11.23 & 57.06 & 9.12 & 103.21 \\
(b) ZS, optimized prompts & 23.42 & 41.04 & 7.14 & 95.47 \\
(c) Best ZS per domain & 23.42 & 57.06 & 9.12 & 119.47 \\
(d) (c) + FT D3 & 23.42 & 57.06 & 33.78 & 152.35 \\
(e) (d) + lower threshold & 23.42 & 57.06 & 36.08 & \textbf{155.42} \\
\bottomrule
\end{tabular}
\caption{Ablation study showing progressive improvements from prompt optimization and fine-tuning.}
\label{tab:ablation}
\end{table}

\begin{figure}[!htbp]
\centering
\def\figPath{teams/team21_QiFans/images/ablation.pdf}
\IfFileExists{\figPath}{
  \includegraphics[width=1.\linewidth]{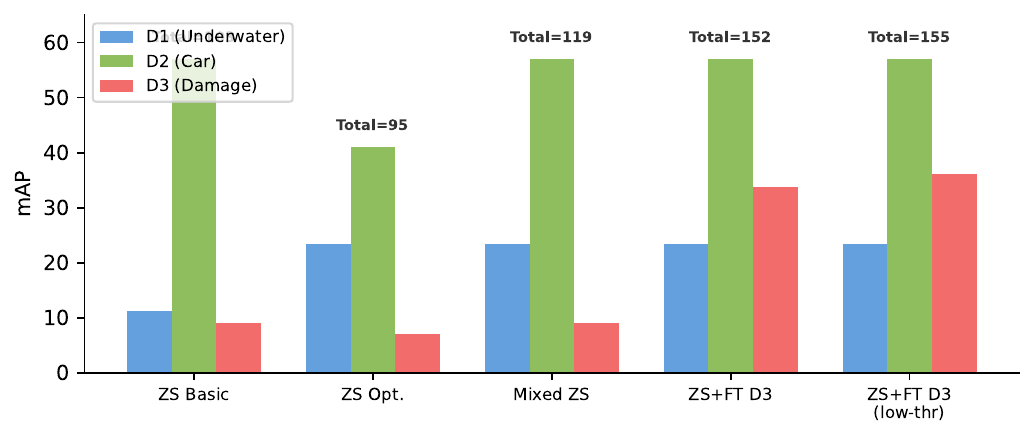}
}{
  \fbox{\parbox[c][0.22\textheight][c]{1.\linewidth}{
  \centering \textbf{Ablation Figure Placeholder}\\
  Put \texttt{ablation.pdf} (or .png/.jpg) in the \texttt{images/} folder (relative to main .tex file) and check filename.
  }}
}
\caption{Ablation results. Prompt optimization boosts D1 (+12 mAP), fine-tuning dramatically improves D3 (+27 mAP), while D2 is near-optimal with simple zero-shot detection.}
\label{fig:ablation}
\end{figure}

\noindent Key findings from their ablation study:
\begin{itemize}
    \item \textbf{Prompt engineering for specialized domains:} Descriptive multi-synonym prompts doubled D1 mAP (11.23 $\to$ 23.42). However, adding synonyms to common objects \emph{hurts} D2 (57.06 $\to$ 41.04) due to spurious false positives.
    \item \textbf{Fine-tuning for fine-grained domains:} D3 improved from 9.12 to 36.08 mAP (+26.96) through fine-tuning on just 60 images, demonstrating the critical importance of domain adaptation for specialized damage categories.
    \item \textbf{Lower thresholds improve recall:} Reducing box\_threshold from 0.2 to 0.1 for D3 improved mAP from 33.78 to 36.08.
    \item \textbf{Failure case:} Self-training with pseudo-labels on support images degraded performance (126.55), likely due to noise in pseudo-labels for fine-grained damage types.
\end{itemize}

\subsection{AIRCAS MILab}

\subsubsection{Proposed Method}
\begin{figure}[!htbp]
\centering
\def\figPath{teams/team11_AIRCAS_MILab/method_figure.pdf}
\IfFileExists{\figPath}{
  \includegraphics[width=0.95\linewidth]{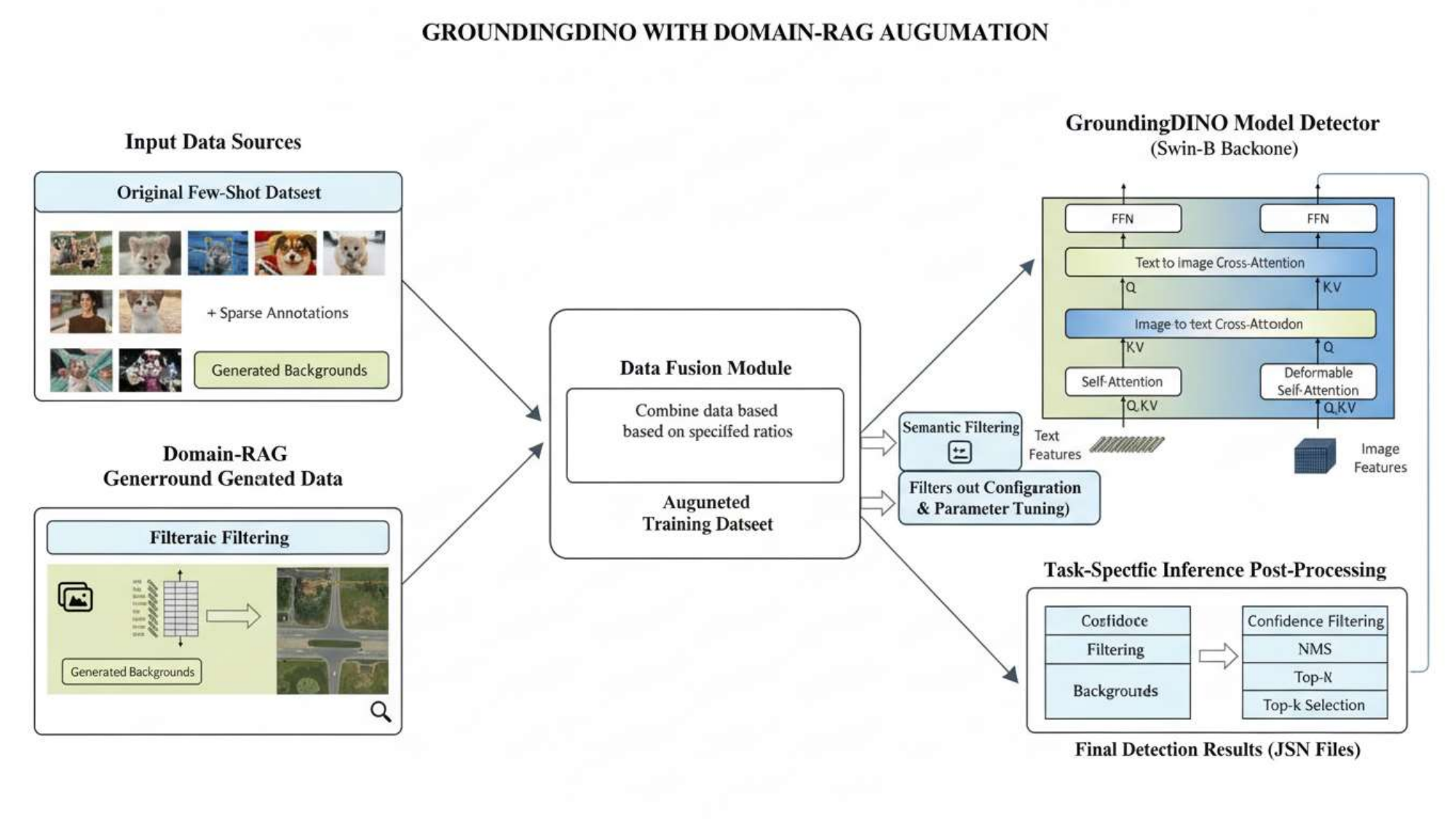}
}{
  \fbox{\parbox[c][0.22\textheight][c]{1.\linewidth}{
  \centering \textbf{Method Figure Placeholder}\\
  Put the method figure file (e.g., .pdf/.png/.jpg) in \texttt{teams/team11_AIRCAS MILab/} folder and check filename.
  }}
}
\caption{Overall pipeline of the AIRCAS MILab team's method. Each dataset-shot pair is treated as an independent few-shot task. Support annotations are expanded with semantically filtered Domain-RAG generated samples, used to build task-specific GroundingDINO configurations, and detector outputs are filtered with dataset-specific post-processing for submission.}
\label{fig:method11}
\end{figure}

The AIRCAS MILab team addresses the cross-domain few-shot object detection challenge by treating it as nine independent few-shot tasks (three target datasets under 1-shot, 5-shot, and 10-shot settings), with the overall pipeline illustrated in Fig.~\ref{fig:method11}.

The AIRCAS MILab team adopts GroundingDINO with a Swin-B visual backbone as the base detector, and fine-tunes it separately for each dataset-shot pair instead of merging all tasks into a single training set. To mitigate the lack of supervision in few-shot scenarios, the team expands original support annotations with Domain-RAG generated samples, and introduces a lightweight semantic filtering step to retain only generated backgrounds semantically compatible with foreground objects, reducing augmentation noise.

Additionally, the team develops an automatic configuration generation and parameter tuning pipeline for GroundingDINO, enabling consistent and reproducible management of all nine task-specific settings (including data paths, class reconstruction, training hyperparameters, checkpoint handling, and inference settings). During inference, dataset-specific post-processing (confidence filtering, NMS, top-$k$ control) is applied to generate final submission files.

The model architecture remains largely consistent with the original GroundingDINO, initialized from official pre-trained checkpoints with no heavy architectural modifications. The focus is on optimizing the data pipeline and task adaptation process: class definitions are automatically reconstructed from target few-shot COCO annotations, and the detector head is adapted to each specific task to ensure stable and robust adaptation to the challenge setting.

The key contributions of the AIRCAS MILab team are as follows:
\begin{itemize}[leftmargin=1.2em]
    \item Semantic filtering is introduced for Domain-RAG generated backgrounds, reducing semantically mismatched augmentation noise before few-shot detector training.
    \item An automatic GroundingDINO configuration generation and parameter tuning pipeline is developed for all nine dataset-shot tasks, covering data paths, class reconstruction, training hyperparameters, checkpoint handling, and inference settings.
    \item Fully task-specific training and dataset-specific inference post-processing are employed (instead of a unified setting), improving stability under cross-domain and extremely low-shot conditions.
\end{itemize}

\subsubsection{Training Details}
The AIRCAS MILab team organizes training around nine independent tasks: dataset1, dataset2, and dataset3 under 1-shot, 5-shot, and 10-shot settings, with no mixing of different datasets or shot settings in a single training set. Each task uses its own task-specific COCO annotation file and GroundingDINO configuration to avoid cross-dataset contamination and preserve the integrity of the few-shot setting.

Original support annotations are converted to task-specific COCO training files and expanded with Domain-RAG generated samples (after semantic filtering to remove semantically inconsistent backgrounds). GroundingDINO is fine-tuned with the AdamW optimizer, and the learning rate and epoch number are adjusted according to the shot setting and target dataset. The validation stage is disabled during training (due to the lack of bounding-box ground truth in official test annotations), with checkpoints saved directly to reduce unnecessary computation.

The team also explored a second-stage pseudo-label re-training scheme (using high-confidence predictions on unlabeled images to augment annotations), but abandoned it in the final submission due to overfitting risks in extremely low-shot conditions. Training and inference are conducted on a single GPU (exact hardware details to be provided by the team).

For inference, the team performs task-specific checkpoint selection and dataset-specific post-processing (instead of global settings). Key post-processing steps include confidence thresholding, per-class NMS, image-id alignment, per-image top-$k$ filtering, and json schema normalization (to match challenge evaluation format). The confidence threshold, NMS IoU threshold, and top-$k$ values are tuned separately for each dataset to adapt to different domain characteristics (background complexity, object scale, false-positive patterns). No ensemble or test-time augmentation (TTA) is used in the final submission.
\subsection{J\_G\_team}

\subsubsection{Proposed Method}
\begin{figure}[!htbp]
\centering
\def\figPath{teams/team12_J_G_team/images/fig1.png}
\IfFileExists{\figPath}{
  \includegraphics[width=1.\linewidth]{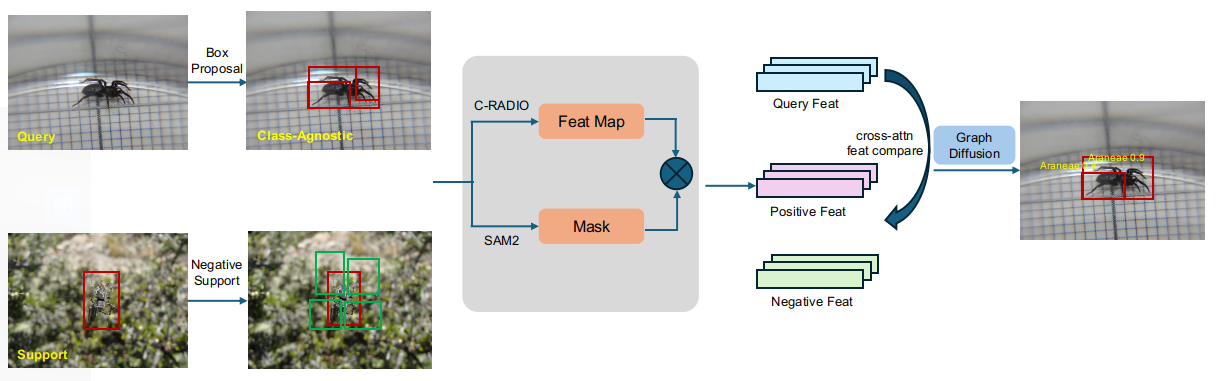}
}{
  \fbox{\parbox[c][0.22\textheight][c]{0.95\linewidth}{
  \centering \textbf{Method Figure Placeholder}\\
  Put \texttt{fig1.png} (or .pdf/.jpg) in \texttt{teams/team12_J_G_team/images/} and check filename.
  }}
}
\caption{Overall pipeline of Negative Prompting for Few-Shot Object Detection.}
\label{fig:method12}
\end{figure}

The J\_G\_team proposes a few-shot object detection method based on FSOD-VFM~\cite{feng2025fsodvfm} with two key improvements to enhance detection accuracy. As shown in Fig.~\ref{fig:method12}, the overall pipeline of this method focuses on strengthening the model's discrimination ability between foreground and background, as well as its adaptability to intra-class variation.

First, the J\_G\_team incorporates negative sample features generated by randomly shifting and scaling positive bounding boxes, which provides the model with additional background context to help distinguish foreground from background. Second, the team replaces static class prototypes with query-adaptive cross-attention prototypes, allowing support features to be dynamically weighted according to each query proposal and thus improving matching precision.

The model architecture of the proposed method consists of three core components: (1) Universal Proposal Network~\cite{jiang2024chatrextamingmultimodalllm} is used as the detector to generate category‑agnostic bounding box proposals for candidate objects; (2) SAM2 is employed as the mask extraction module to produce accurate object masks for support annotations, aiding robust feature extraction; (3) C-RADIOv4-H is adopted as the visual feature extractor for the whole image. During inference, the J\_G\_team sets specific preprocessing rules: C-RADIOv4 resizes all input images so that the longer side is 1260, and for all proposals generated by UPN, the top 100 boxes per image ranked by confidence are selected.

\subsubsection{Training Details}
The method proposed by the J\_G\_team is training-free and does not require any task-specific finetuning, which is a key characteristic of its implementation process.

\subsection{NTR}
\subsubsection{Proposed Method}

\noindent \textbf{Problem Formulation.} Cross-Domain Few-Shot Object Detection (CD-FSOD) aims to detect novel object categories in a target domain given only $K$ annotated support examples per category, where the target domain differs significantly from the source domain.

Formally, let $\mathcal{S} = \{(I_i^s, B_i^s, c_i)\}_{i=1}^{NK}$ denote the support set of $N$ categories with $K$ shots each, where $I_i^s$ is a support image, $B_i^s$ is its bounding box annotation, and $c_i$ is the category label. Given a query image $I^q$ and support set $\mathcal{S}$, the model must predict a set of bounding boxes $\hat{B} = \{(b_j, c_j, s_j)\}$ on $I^q$, where $b_j$, $c_j$, and $s_j$ denote the box coordinates, predicted category, and confidence score, respectively.

The challenge evaluates models under $K \in \{1, 5, 10\}$ shots across three novel test domains (marine fauna, car detection, and car damage detection) using mean Average Precision (mAP). The final ranking score weights 1-shot performance more heavily:
\begin{equation}
\text{Score} = 2 \cdot \overline{\text{mAP}}_{1\text{-shot}} + \overline{\text{mAP}}_{5\text{-shot}} + \overline{\text{mAP}}_{10\text{-shot}},
\end{equation}
where $\overline{\text{mAP}}_{K\text{-shot}}$ is the mean mAP across the three datasets under $K$-shot setting.

\noindent \textbf{Overview: Training-Free FSOD-VFM}

Theyadopt \textbf{FSOD-VFM}~\cite{feng2026fsodvfm}, a training-free few-shot object detection framework that integrates three vision foundation models:

\begin{enumerate}[noitemsep]
    \item \textbf{UPN}~\cite{zhao2024upn} --- Universal Proposal Network for category-agnostic proposal generation;
    \item \textbf{SAM2}~\cite{ravi2024sam2} --- Segment Anything Model 2 for precise mask extraction;
    \item \textbf{DINOv2}~\cite{oquab2024dinov2} ViT-L/14 --- for discriminative feature extraction.
\end{enumerate}

No fine-tuning is performed on the target domain. The method uses only the provided $K$-shot support set at test time.

\noindent \textbf{Method Details}

The FSOD-VFM pipeline proceeds in three stages:

\paragraph{Stage 1: Proposal Generation.}
Given a query image $I^q$, UPN generates a set of category-agnostic bounding box proposals $\mathcal{P} = \{p_1, \ldots, p_M\}$. UPN is pretrained on large-scale data and generalizes to novel domains without retraining.

\paragraph{Stage 2: Feature Matching.}
DINOv2 ViT-L/14 extracts patch-level features from both the support crops and the query proposals. For each support example $(I_i^s, B_i^s)$, the region within $B_i^s$ is cropped and encoded. Query proposal features are matched against support features via cosine similarity to assign category scores.

To address over-fragmentation in UPN proposals (many small, redundant boxes), FSOD-VFM introduces a \textbf{graph-based confidence reweighting} strategy. A graph $G = (V, E)$ is constructed over proposals, where edges connect spatially overlapping proposals. Confidence scores are diffused along the graph to propagate information from high-confidence to neighboring proposals, suppressing spurious detections while preserving true positives.

\paragraph{Stage 3: SAM2 Mask Refinement.}
SAM2 refines proposal boundaries by predicting instance masks for top-scored proposals. The tighter mask-derived bounding boxes improve localization accuracy.

Final detections are produced by non-maximum suppression over the refined, reweighted proposals.

\paragraph{Hyperparameters.}
Theyuse the default FSOD-VFM hyperparameters: diffusion steps $= 30$, graph diffusion weight $\alpha = 0.3$, NMS threshold $\lambda = 0.5$, minimum confidence threshold $= 0.01$.

\noindent \textbf{Foundation Models Used}

Per competition rules, theyuse exactly three foundation models:
\begin{enumerate}[noitemsep]
    \item DINOv2 ViT-L/14~\cite{oquab2024dinov2} (pretrained on LVD-142M, publicly available)
    \item SAM2.1 large~\cite{ravi2024sam2} (pretrained by Meta AI, publicly available)
    \item UPN large~\cite{zhao2024upn} (pretrained for universal proposal generation, publicly available)
\end{enumerate}

\noindent No additional datasets or training data beyond the provided $K$-shot support sets are used.





\subsubsection{Training Details}

\textbf{Platform:} NVIDIA H100 (80 GB) GPU. PyTorch 2.6.0, Python 3.10, CUDA 12.4.

\textbf{Inference runtime:} Approximately 30--90 minutes per dataset/shot case on a single NVIDIA H100 GPU, depending on the number of query images. Total runtime for all 9 cases: approximately 8 hours.

\textbf{Extra training data:} None. This is a training-free method. No additional datasets beyond the provided $K$-shot support sets are used at any stage.

\textbf{Pretrained models:} DINOv2 ViT-L/14, SAM2.1 large, and UPN large (all publicly available). No fine-tuning is applied to any foundation model.

\textbf{Code:} Available in the accompanying code submission (\texttt{NTR\_code.zip}). The code is based on the publicly available FSOD-VFM repository~\cite{feng2026fsodvfm}. Setup requires compiling the UPN CUDA extension (MultiScaleDeformableAttention) and installing SAM2.

\textbf{Challenge impressions:} The CD-FSOD challenge is a well-designed benchmark that highlights a critical limitation of standard FSOD methods --- they often overfit to the source domain and fail when the target domain introduces significant visual style or class distribution shifts. They found that training-free vision foundation models generalize surprisingly well across diverse unseen domains (marine fauna, vehicles, damage detection), outperforming a fine-tuned GroundingDINO baseline on test domains despite the fine-tuned model performing better on validation domains. This suggests that over-parameterized few-shot adaptation can hurt cross-domain generalization. The challenge's emphasis on 1-shot performance (2$\times$ weight) correctly prioritizes the most challenging and practically relevant setting.



\subsection{WRC}

\subsubsection{Proposed Method}
\begin{figure}[!htbp]
\centering
\def\figPath{teams/team13_WRC/images/method_figure.png}
\IfFileExists{\figPath}{
  \includegraphics[width=0.95\linewidth]{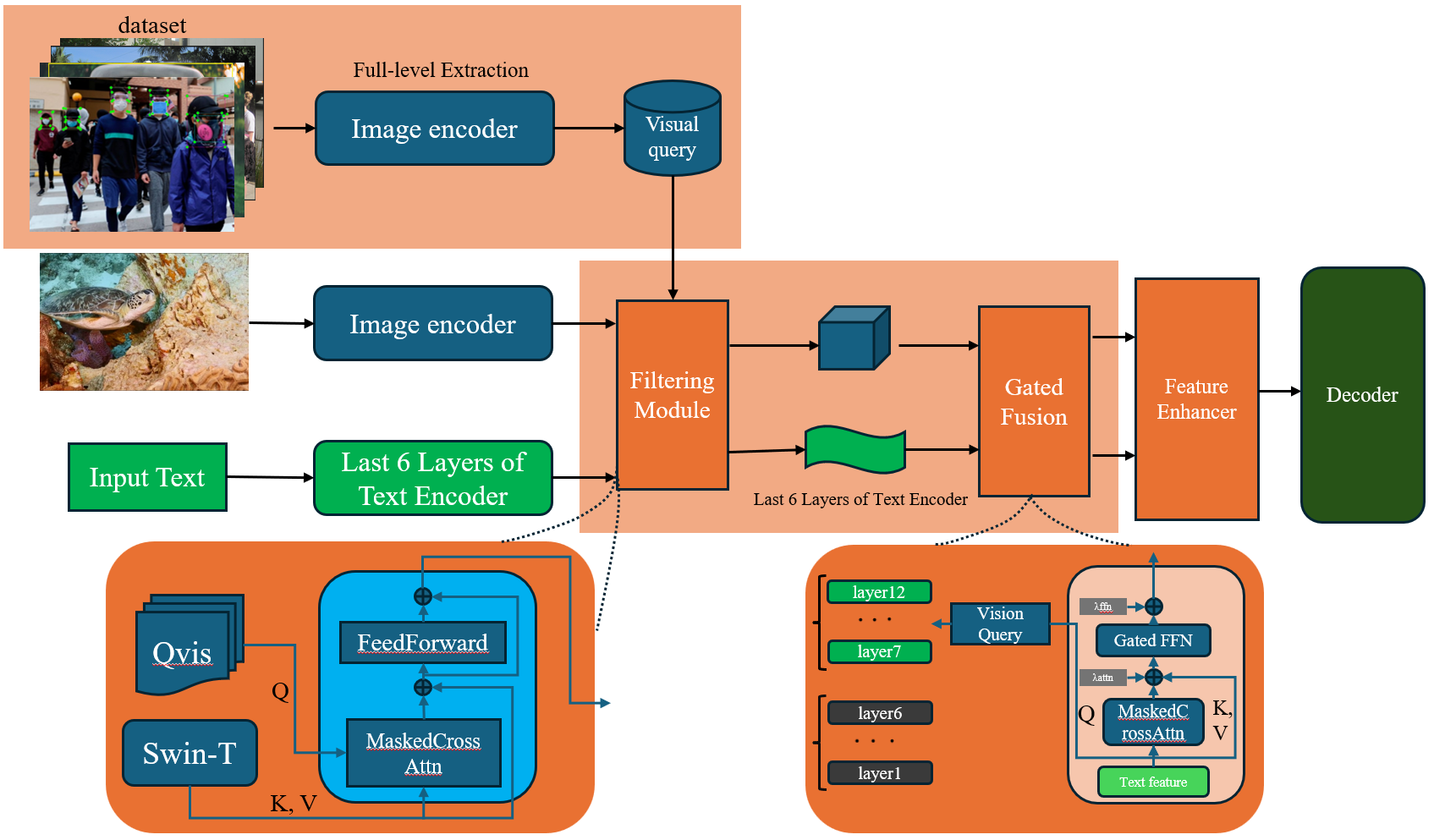}
}{
  \fbox{\parbox[c][0.8\textheight][c]{1.\linewidth}{
  \centering \textbf{Method Figure Placeholder}\\
  Put \texttt{method\_figure.png} (or .png/.jpg) in \texttt{teams2026/team13_WRC/images/} folder and check filename.
  }}
}
\caption{Overall pipeline of Multimodal Query-based Object Detection.}
\label{fig:method13}
\end{figure}

Based on the image-text multimodal detector GroundingDINO, the WRC team proposes a novel framework for multimodal query-based object detection, whose overall pipeline is illustrated in Fig.~\ref{fig:method13}.

At a high level, the WRC team first constructs a full-level multi-scale visual query library, and performs instantiation and cascaded refinement on retrieved visual features under the guidance of text semantics. The refined visual queries are then embedded into the deep layers of the frozen text encoder through a gated adaptive fusion mechanism, realizing efficient cross-modal alignment between visual and textual features.

For model training, the team conducts joint image-text fine-tuning with a lightweight strategy: the visual backbone network and shallow layers of the text encoder are frozen to retain basic representation ability. Learnable continuous prompt parameters are introduced into the text branch, and LoRA low-rank adaptation is applied to the cross-attention layer for visual-text fusion. To handle partially unlabeled instances in few-shot datasets, the team incorporates the EMA Mean Teacher to generate pseudo-labels, which strengthens fine-tuning performance. Only the parameters of the fusion module are tuned to improve few-shot object detection performance.

In terms of model architecture, the WRC team uses Swin-T as the visual backbone and BERT-base-uncased as the text encoder, with Deformable DETR as the detection framework. Additional key components include a Vision Query Bank, gated cross-attention, CoOp learnable prompts, and LoRA modules, which are integrated into the vision-language fusion pipeline to bridge the modal gap and align fine-grained visual-text features.

The key contributions of the WRC team are summarized as follows:
\begin{itemize}
    \item A vision-enhanced text branch is built to reduce the modality gap of pure text in describing fine-grained visual characteristics, through constructing a visual query library, cascaded feature refinement, and gated adaptive fusion for efficient cross-modal alignment.
    \item A lightweight joint image-text fine-tuning scheme is designed, which freezes the visual backbone and shallow text encoder, applies learnable prompts and LoRA adaptation, and only fine-tunes the fusion module to improve few-shot detection performance and alleviate overfitting.
\end{itemize}

\subsubsection{Training Details}
The WRC team conducts experiments on the COCO17 dataset under few-shot settings of 1/5/10-shots, using fixed support sets provided by the challenge. The model is optimized with the AdamW optimizer at a learning rate of 1e-4, equipped with a WarmupMultiStepLR scheduler. Training runs for 20 epochs with 1 warm-up epoch.

Data augmentations applied during training include multi-scale resizing, random horizontal flipping, and standard normalization. All experiments are run on a single NVIDIA A100 GPU, with a total training time of approximately 3 hours.

For inference, the model uses a fixed test scale with a short edge of 800 and a maximum long edge of 1333, a score threshold of 0.05, and a maximum number of detections set to 900. No test-time augmentation, model ensemble, or extra post-processing operations are adopted during the inference stage.

\subsection{NUDT-RSIP}
\subsubsection{Method overview}

\begin{figure}[!htbp]
\centering
\def\figPath{teams/team7_NUDT-RSIP/images/NTIRE-ao.pdf}
\IfFileExists{\figPath}{
  \includegraphics[width=1.\linewidth]{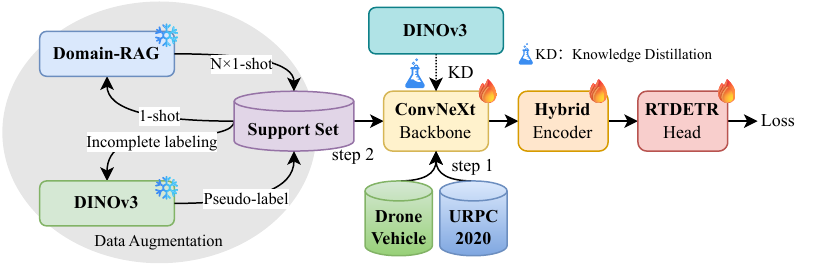}
}{
  \fbox{\parbox[c][0.22\textheight][c]{0.95\linewidth}{
  \centering \textbf{Method Figure Placeholder}\\
  Put \texttt{fig1.png} (or .pdf/.jpg) in \texttt{teams/team12_J_G_team/images/} and check filename.
  }}
}
\caption{Overall pipeline of \texttt{DDR}.}
\label{fig:method144}
\end{figure}

\noindent \textbf{High-level idea:} As illustrated in Fig.~\ref{fig:method144}, they propose a synergistic strategy integrating data augmentation, model distillation, and two-stage fine-tuning to systematically address data scarcity and domain shift in cross-domain few-shot object detection. At the data level, they employ Domain-RAG to enrich the diversity of the support set and leverage a powerful DINOv3 model to re-label unlabeled data, generating high-quality pseudo-labels to enhance supervision. For the model architecture, they adopt DINOv3-ConvNeXt-Large as the feature extraction backbone combined with an RT-DETR head, balancing strong representational capacity with improved detection accuracy. Finally, a two-stage fine-tuning scheme is designed: the pretrained model is first adapted on Drone Vehicle and URPC2020 datasets, followed by target-specific fine-tuning on the augmented support set. This multi-level framework enables robust generalization in complex cross-domain scenarios under extremely limited annotation budgets.\\
\noindent \textbf{Key contributions:}
\begin{itemize}
    \item \textbf{Contribution 1:} They propose a unified data enhancement paradigm that combines Domain-RAG for support set diversification with DINOv3-driven pseudo-labeling to mitigate annotation sparsity in few-shot learning.
    \item \textbf{Contribution 2:}  They introduce a high-performance DINOv3-ConvNeXt-Large feature extractor with knowledge distillation to transfer robust representations while adapting to novel target domains.
    \item \textbf{Contribution 3:} They design a two-stage progressive fine-tuning strategy that first adapts on Drone Vehicle and URPC2020 datasets before target-specific fine-tuning, effectively bridging the domain gap under limited annotations.
\end{itemize}

\subsubsection{Model architecture}
Describe the backbone, detector head, any added modules, and where they are inserted.
\begin{itemize}
    \item Backbone: DINOv3-ConvNeXt-Large
    \item Detector: RT-DETR
    \item Additional components: Domain-RAG
\end{itemize}

\subsection{French Borelli}
\subsubsection{Proposed Method}
\begin{figure}[!htbp]
\centering
\def\figPath{teams/team23\_French\_Borelli/method_figure.png}
\IfFileExists{\figPath}{
  \includegraphics[width=1.\linewidth,trim=30 10 100 10,clip]{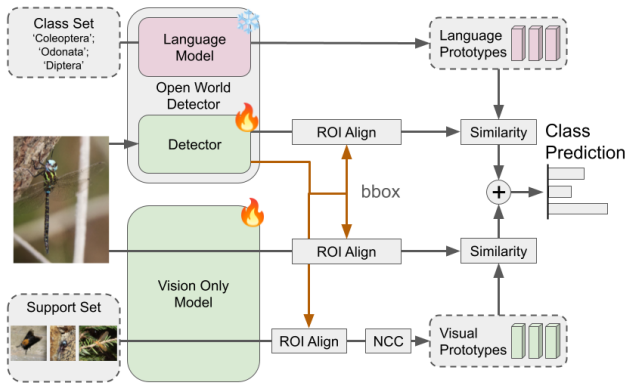}
}{
  \fbox{\parbox[c][0.22\textheight][c]{1.\linewidth}{
  \centering \textbf{Method Figure Placeholder}\\
  Put \texttt{method\_figure.png} (or .pdf/.jpg) in \texttt{teams2026/team23\_French\_Borelli/} and check filename.
  }}
}
\caption{Overall pipeline of Triple-Tower. 
They start from a double-tower open-vocabulary detector composed of a language tower and a vision tower. The language tower encodes class names into language prototypes, while the vision tower extracts image features and predicts object bounding boxes. They introduce a third uni-modal tower, a vision-only model trained with a nearest centroid classifier (NCC), which builds visual prototypes from the support set. 
}
\label{fig:method23}
\end{figure}

\noindent \textbf{High-level idea:} 
Open-vocabulary detectors are designed for zero-shot generalization, but there is no straightforward way to improve their performance in few-shot settings beyond standard fine-tuning.
In this work, They argue that few-shot adaptation benefits from decoupling localization from classification. The detector already localizes well, while classification can be improved with a vision-only auxiliary model that exploits richer visual structure than language-aligned features. The model architecture is shown in Fig.~\ref{fig:method23}.

Their method starts by fine-tuning the open-vocabulary detector WeDetect~\cite{fu2025wedetect}.
During fine-tuning, they add pseudo annotations, since they observe that unannotated objects in the training set hurt performance by biasing the detector toward predicting fewer objects.
Finally, they improve the class prediction of each detected bounding box with a vision-only model, DINOv3~\cite{simeoni2025dinov3}. 
To remain compute-efficient, they perform parameter efficient fine-tuning by only training LayerNorm layers (0.03\% of weights).
\\
\noindent \textbf{Key contributions:}
\begin{itemize}
    \item \textbf{Contribution 1:} \textbf{Vision Boosting}: Using a vision model with parameter efficient fine-tuning to boost few-shot classification performance.
    \item \textbf{Contribution 2:} \textbf{Pseudo Annotations}: Periodically adding high-confidence detector predictions from the support set as pseudo annotations during training. 
\end{itemize}

\noindent \textbf{Model architecture}
\begin{itemize}
    \item Backbone: WeDetect~\cite{fu2025wedetect} (ConvNeXt-base and X-BERT-base) and DINOv3~\cite{simeoni2025dinov3} (ViT-L)
\end{itemize}

\subsubsection{Training details}
\begin{itemize}
    \item Training data: No additional training data
    \item Pseudo Annotations:
    Pseudo annotations are generated during the fine-tuning of WeDetect over 100 epochs. Starting from epoch 15, they run the detector on the support set every 5 epochs and add high-confidence predictions to the training annotations. Predictions whose score is above a fixed threshold are kept, new detections with excessive overlap (IoU$\geq$0.70) with existing support annotations are discarded.
    \item Vision Boosting:
    After fine-tuning the open-vocabulary detector, they train the vision backbone with an Nearest Centroid Classifier (NCC) loss on all labels and pseudo labels.
    \item Optimization:
    For WeDetect: they optimize the detector with AdamW, using a learning rate of \texttt{2e-5}, weight decay \texttt{0.05}, batch size \texttt{4}, and a total of \texttt{100} training epochs. 
    For DINOv3: they optimize the detector with AdamW, using a learning rate of \texttt{0.005} over \texttt{20} epochs with batch size \texttt{50}
    \item Augmentations:
    Training uses random horizontal flip, multi-scale resizing, random cropping, YOLO-style mosaic augmentation, random affine transformations, MixUp, HSV augmentation, blur, median blur, grayscale conversion, and CLAHE. 
    For aerial datasets (dataset2), they add rotation augmentation.
    \item Hardware / runtime:
    All experiments were conducted on a single NVIDIA GeForce RTX 3090. Training over all datasets takes approximately 12min in total, with 10min for the detector and 2min for the vision models.
\end{itemize}

\noindent \textbf{Inference details}
\begin{itemize}
    \item Test-time settings:
     Images are resized to \texttt{640$\times$640} using keep-ratio resizing followed by letterbox padding. During inference, They use multi-label prediction, a score threshold of \texttt{0.0001}, pre-NMS filtering with \texttt{nms\_pre=30000}, non-maximum suppression with IoU threshold \texttt{0.7}, and keep at most \texttt{200} detections per image.
    \item Ensemble:  At inference, the final classification logits are obtained by adding the predictions of the vision-based classifier and the language-based classifier.
\end{itemize}

\section{Special Closed-Source Track Methods}
\label{sec:teams-solution2}
\subsection{FewShotEverything}

\subsubsection{Proposed Method}
\begin{figure}[!htbp]
\centering
\def\figPath{teams/team24_FewShotEverything/images/method.pdf}
\IfFileExists{\figPath}{
  \includegraphics[width=\linewidth]{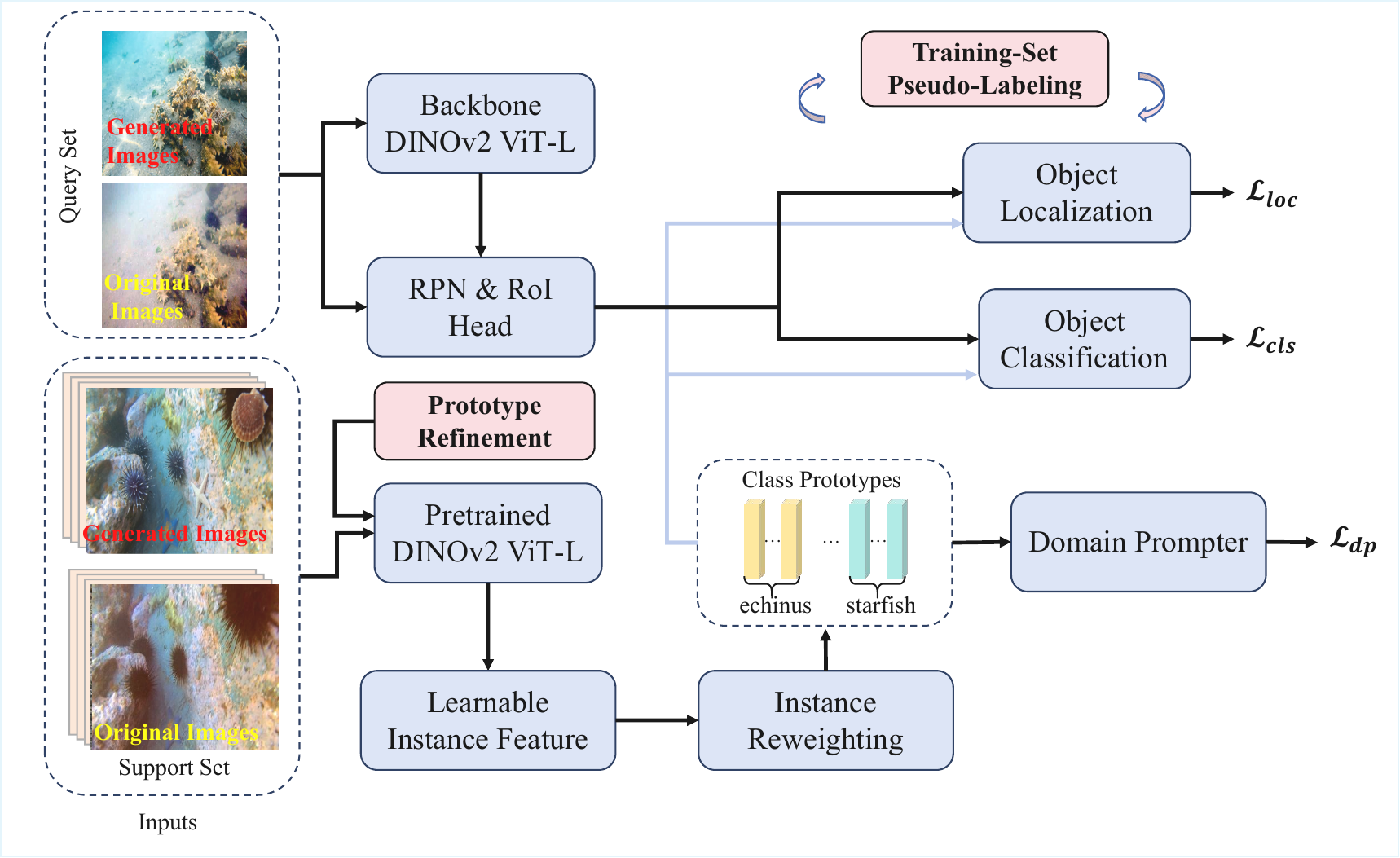}
}{
  \fbox{\parbox[c][0.22\textheight][c]{\linewidth}{
  \centering \textbf{Method Figure Placeholder}\\
  Put \texttt{method.pdf} (or .png/.jpg) in \texttt{teams/team24_FewShotEverything/images/} and check filename.
  }}
}
\caption{Overall pipeline of AIPR.}
\label{fig:method}
\end{figure}


\begin{figure}[!htbp]
\centering
\def\figPath{teams/team24_FewShotEverything/images/generation.pdf}
\IfFileExists{\figPath}{
  \includegraphics[width=\linewidth]{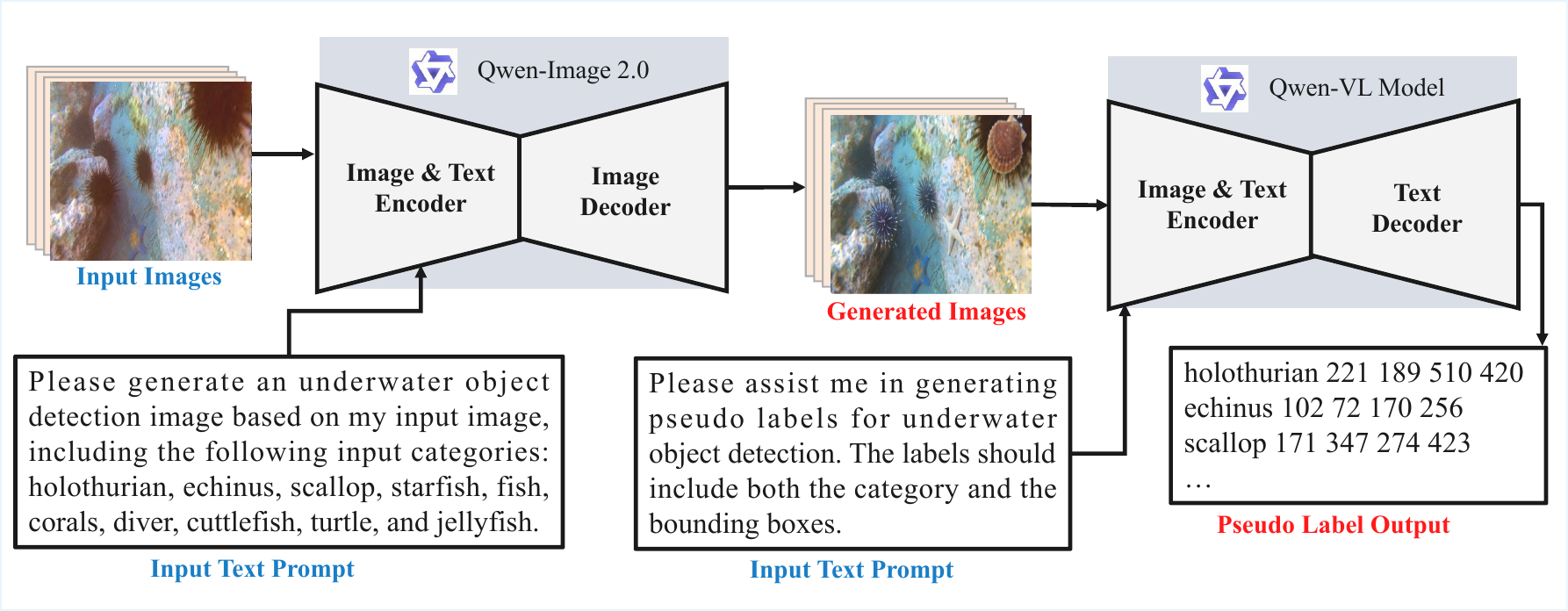}
}{
  \fbox{\parbox[c][0.22\textheight][c]{\linewidth}{
  \centering \textbf{Data Generation Figure Placeholder}\\
  Put \texttt{generation.pdf} (or .png/.jpg) in \texttt{teams/team24_FewShotEverything/images/} and check filename.
  }}
}
\caption{Overall pipeline of the data generation process for data augmentation.}
\label{fig:generation}
\end{figure}

\noindent \textbf{High-level idea:} They proposed a Cross-Domain Few-Shot Object Detection (CDFSOD) framework based on the CD-ViTO baseline\cite{fu2024cross} under the closed-setting.
The framework consists of three modules: (1) \textbf{Training-Set Data Augmentation module (TDAM)} that enhances sample diversity and exposes the model to more varied object appearances in target domain, which improves robustness to domain shift between the source and target domains.
(2) \textbf{Prototype Refinement Module (PRM)} that refines foreground prototype extraction by concentrating on semantically relevant object regions, which effectively mitigates the influence of surrounding noise and irrelevant background, leading to more reliable cross-domain prototypes.
(3) \textbf{Iterative Pseudo-Labeling Module (IPLM)}, which addresses incomplete support annotations by progressively identifying unlabeled target instances in support images. Specifically, it identifies cases where only a single instance is labeled per image despite the presence of multiple targets. To prevent the model from treating unlabeled vehicles as background in dataset2, the supervision is enriched through iterative pseudo-label refinement.
Together, these modules improve detection performance and effectively reduce the impact of domain shift.

\noindent \textbf{Key contributions:}
\begin{itemize}
  \item \textbf{Contribution 1:} A Training-Set Data Augmentation module is proposed by leveraging a VLM to synthesize support-like images and corresponding detection annotations, enhancing sample diversity and improving robustness to domain shift between the source and targets.
    \item \textbf{Contribution 2:} A Prototype Refinement Module is introduced to improve foreground prototype extraction by focusing on semantically relevant object regions, thereby reducing the interference of background noise and producing more reliable cross-domain prototypes.
    \item \textbf{Contribution 3 (optional):} An Iterative Pseudo-Labeling Module is designed to address incomplete support annotations by progressively mining unlabeled target instances and refining pseudo-labels, which provides additional supervision for target-domain adaptation.
\end{itemize}

\noindent\textbf{Model architecture.}
Describe the backbone, detector head, any added modules, and where they are inserted.
\begin{itemize}
    \item Backbone: Following baseline work \cite{fu2024cross}, they utilize the pre-trained ViT-L/14 variant from DINOv2 as backbone, which is initialized with weights learned through self-supervised learning on large-scale image datasets. This backbone extracts high-level semantic features from input images via its multi-layer transformer encoder.
    \item Detector:  They then integrate the backbone with the Faster R-CNN~\cite{fasterrcnn} architecture, which consists of a Region Proposal Network (RPN) and a detection head for classification and bounding box regression. 
    \item Additional components: Following baseline work \cite{fu2024cross}, they utilize the pre-trained ViT-L/14 variant from DINOv2 as the backbone for prototype extraction, leveraging its powerful feature representation capabilities to enhance the accuracy of CD-FSOD.
\end{itemize}

\subsubsection{Training details}
Provide training settings sufficient for reproduction.
\begin{itemize}
    \item Training data: Following closed-source setting and CD-ViTO~\cite{fu2024cross}, COCO~\cite{lin2014microsoft} is used as the source-domain dataset for base training, since its large-scale annotations and diverse object categories provide strong transferable detection features. The three datasets provided by the challenge are treated as the target-domain data.
    \item Few-shot setting: Following provided annotation splits of the targets, they evaluate under 1-shot, 5-shot, and 10-shot settings, where each class is assigned 1, 5, or 10 annotated support instances, respectively, and only the support annotations corresponding to each shot setting are used for model adaptation and training.
    \item Optimization: The Training-Set Data Augmentation module and Prototype Refinement Module are used for all three datasets, while the Iterative Pseudo-Labeling Module is additionally applied to Dataset2, which is a single-class dataset with incomplete training annotations. 
    The trainable parameters are fine-tuned with dataset-specific training epochs under different experimental settings. For all datasets, the model is optimized using SGD with a base learning rate of 0.001, and the batch size is set to 16 images per batch.
    \item Augmentations: The target-domain training set is augmented via Qwen-Image2.0\cite{wu2025qwen}. Given a target-domain image and a designed text prompt, Qwen-Image2.0 generates additional images that are visually similar, while preserving domain characteristics and target objects. These synthesized images, along with the corresponding prompts, are then fed into a Qwen-VL model\cite{wang2024qwen2} to produce pseudo-labels, including object categories and bounding boxes.
    \item Hardware / runtime: Experiments are conducted on two NVIDIA A6000 GPUs.
\end{itemize}

\noindent\textbf{Inference details}
Explain inference-time details that affect performance.
\begin{itemize}
    \item Test-time settings: they follow the default inference
pipeline of the corresponding detector framework for
prediction filtering and box selection.
    \item Ensemble / TTA: None
    \item Post-processing: In the iterative pseudo-labeling module, low-confidence pseudo labels generated from initial training data are removed based on a confidence threshold.
\end{itemize}

\subsection{Fusion-Few}
\subsubsection{Proposed Method}


\begin{figure}[!htbp]
\centering
\def\figPath{teams/team2_Fusion-Few/method_figure.pdf}
\IfFileExists{\figPath}{
  \includegraphics[width=0.95\linewidth]{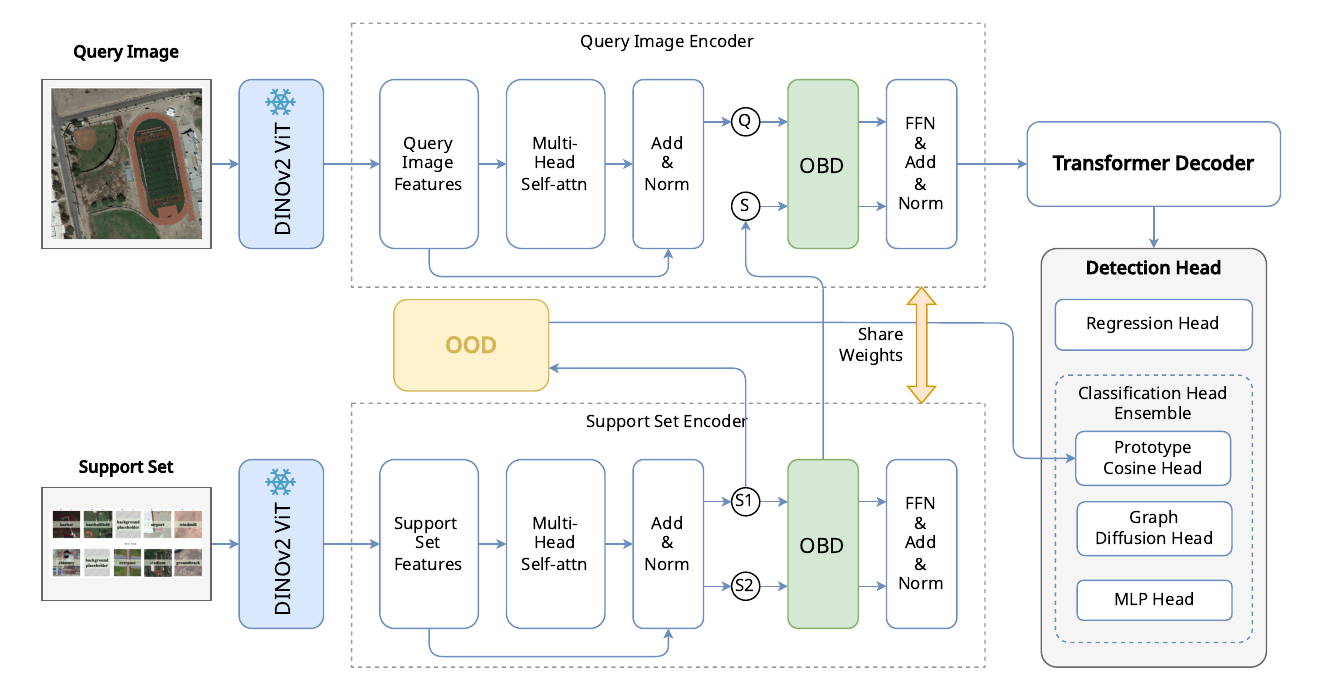}
}{
  \fbox{\parbox[c][0.22\textheight][c]{1.\linewidth}{
  \centering \textbf{Method Figure Placeholder}\\
  Put \texttt{method\_figure.pdf} (or .png/.jpg) in \texttt{teams/team2_Fusion-Few/} and check filename.
  }}
}

\caption{Overview of the FusionFormer framework. The architecture leverages a frozen DINOv2 ViT backbone for feature extraction. It features a Siamese-style encoder-decoder structure integrated with Object-Background Discrimination (OBD) and Object-Object Discrimination (OOD) modules to mitigate cross-domain feature confusion. The final detection head employs a multi-strategy ensemble (Prototype Cosine, Graph Diffusion, and MLP) to enhance classification robustness in few-shot scenarios.}
\label{fig:method2}
\end{figure}

\noindent \textbf{High-level idea:} They present FusionFormer, a Transformer-based architecture specifically engineered for Cross-Domain Few-Shot Object Detection (CD-FSOD). The core objective is to mitigate the severe feature confusion inherent in cross-domain scenarios. Built upon the CDFormer~\cite{meng2025cdformer} backbone, their method introduces a learnable background token within the Object-Background Discrimination (OBD) module and contrastive learning within the Object-Object Discrimination (OOD) module to enhance background separation and inter-class distinctness. To further improve inference robustness, they propose an offline ensemble strategy that synergistically fuses predictions from a linear classifier, a graph diffusion model, and OOD-based feature caching. Combined with reinforced source-domain augmentation, test-time pseudo-labeling, and Test-Time Augmentation (TTA), FusionFormer effectively bridges the domain gap and exhibits superior generalization on unseen target domains.\\
\noindent \textbf{Key contributions:}
\begin{itemize}
    \item \textbf{Enhanced Detection Framework:} A CDFormer~\cite{meng2025cdformer}-based architecture that fundamentally addresses feature confusion in cross-domain settings via the dual-pronged OBD and OOD modules.
    \item \textbf{Innovative Multi-Head Ensemble:} An offline ensemble mechanism that integrates a linear layer, graph diffusion, and feature-cache information to significantly boost prediction accuracy.
    \item \textbf{Comprehensive Optimization Pipeline:} A robust training-to-inference workflow incorporating specialized augmentation, high-confidence pseudo-label mining, and TTA to minimize performance degradation.
\end{itemize}

\noindent \textbf{Model architecture.}
Describe the backbone, detector head, any added modules, and where they are inserted.
\begin{itemize}
    \item Backbone: A frozen DINOv2 ViT-L/14 is used as the main feature extractor. To refine spatial granularity, a convolutional layer is appended to downsample stride-14 features to stride-28, ensuring better alignment for object detection.
    \item Detector: The detector follows a Deformable-DETR architecture, comprising a 6-layer Encoder and a 6-layer Decoder. The decoder is configured as a category-agnostic meta-decoder with 300 queries to maintain flexibility across domains.
    \item Additional components:
    \begin{itemize}
        \item Query-Support Interaction: A SingleHeadSiameseAttention module is integrated into the first encoder layer to facilitate cross-modal feature interaction between query and support branches.
        \item Episodic Category Codes: Category prototypes derived from support branch are injected into the main branch per episode to provide strong class-specific priors.
        \item Multi-type Prediction Heads: the regression head utilizes a standard MLP; the classification task is handled by a triple-head ensemble: a linear layer, a graph diffusion model, and an OOD-based feature cache.
    \end{itemize}
\end{itemize}

\subsection{nudt\_0110Dplter}
\subsubsection{Method overview}

\textbf{High-level idea:} 
Building upon CD-ViTO baseline, they propose two key optimizations to address the challenges of CD-FSOD: 1) They introduce a multi-scale prototype fusion mechanism that constructs image pyramids (scales of 0.9x, 1.0x, 1.1x, and 1.2x) and employs temperature-scaled soft weighting (temperature=0.1) to generate robust prototype features resilient to scale variations. 2) They design an Enhanced Training Strategy (ETS) incorporating RandomFlip, RandomChoiceResize (11 discrete scales), and RandomCrop to significantly expand the diversity of limited support sets, mitigating overfitting. These lightweight optimizations integrate seamlessly with CD-ViTO's learnable instance features and domain prompter, achieving substantial performance gains especially on targets with significant indefinable boundaries (IB).

\textbf{Key contributions:}
\begin{itemize}
    \item Contribution 1: A multi-scale prototype extraction module that fuses DINOv2 features from multiple image scales via learnable soft-weighted aggregation, enhancing scale-invariance without increasing inference cost.
    \item Contribution 2: An Enhanced Training Strategy (ETS) specifically designed for few-shot scenarios, utilizing RandomChoiceResize across 11 scales and RandomCrop to augment extremely limited support sets (1-10 shots).
    \item Contribution 3: A practical optimization framework that maintains the parameter efficiency of CD-ViTO (only +0.8M trainable parameters) while significantly improving cross-domain generalization on challenging underwater and industrial defect datasets.
\end{itemize}

\subsubsection{Model architecture}

Their method builds upon the CD-ViTO (Cross-Domain Vision Transformer) architecture, inheriting its core components including the learnable instance features ($M_{LIF}$), instance reweighting module ($M_{IR}$), and domain prompter ($M_{DP}$). They introduce two key augmentations to the prototype extraction and training pipeline:

\begin{itemize}
    \item \textbf{Backbone:} A frozen DINOv2 ViT-L/14 pre-trained on ImageNet-22K via self-supervised learning.
    \item \textbf{Detector:} DE-ViT based open-set detector with Region Proposal Network (RPN), RoI Align, Detection Head ($M_{DET}$), and One-vs-Rest Classification Head ($M_{CLS}$).
    \item \textbf{Additional components:} 
    \begin{enumerate}
        \item \textit{Multi-Scale Prototype Extraction (MSPE):} During support set processing, they generate an image pyramid with scales [0.9, 1.0, 1.1, 1.2], extract patch tokens from each scale using DINOv2, and fuse them via temperature-scaled softmax weighting (temperature=0.1) to obtain scale-robust prototypes.
        \item \textit{ETS Augmentation Pipeline:} Sequential application of RandomFlip (prob=0.5), RandomChoiceResize (short edge randomly selected from [480, 512, 544, 576, 608, 640, 672, 704, 736, 768, 800], max\_size=1333), and RandomCrop (target size 384$\times$600) applied dynamically to support set images during training.
    \end{enumerate}
\end{itemize}

\section*{Acknowledgments}
This work was partially supported by the Humboldt Foundation. We thank the NTIRE 2026 sponsors: OPPO, Kuaishou, and the University of Wurzburg (Computer Vision Lab).

\clearpage
\appendix

\section{Teams and affiliations}
\label{sec:teams}
\subsection*{NTIRE 2026 team}
\noindent\textit{\textbf{Title: }} NTIRE 2026 Challenge on Cross-Domain Few-Shot Object Detection: Methods and Results.\\
\noindent\textit{\textbf{Members: }} \\
Xingyu Qiu$^{1}$ (\href{mailto:xyqiu24@m.fudan.edu.cn}{xyqiu24@m.fudan.edu.cn}),\\
Yuqian Fu$^{2}$ (\href{mailto:yuqian.fu.ai@gmail.com}{yuqian.fu.ai@gmail.com}),\\
Jiawei Geng$^{1}$ (\href{mailto:jwgeng25@m.fudan.edu.cn}{jwgeng25@m.fudan.edu.cn}),\\
Bin Ren$^{3}$ (\href{mailto:bin.ren.mondo@gmail.com}{bin.ren.mondo@gmail.com}),\\
Jiancheng Pan$^{4}$ (\href{mailto:jiancheng.pan.plus@gmail.com}{jiancheng.pan.plus@gmail.com}),\\
Zongwei Wu$^{5}$ (\href{mailto:zongwei.wu@uni-wuerzburg.de}{zongwei.wu@uni-wuerzburg.de}),\\
Hao Tang$^{6}$ (\href{mailto:howard.haotang@gmail.com}{howard.haotang@gmail.com}),\\
Yanwei Fu$^{1}$ (\href{mailto:yanweifu@fudan.edu.cn}{yanweifu@fudan.edu.cn}),\\
Radu Timofte$^{5}$ (\href{mailto:radu.timofte@uni-wuerzburg.de}{radu.timofte@uni-wuerzburg.de}),\\
Nicu Sebe$^{7}$ (\href{mailto:niculae.sebe@unitn.it}{niculae.sebe@unitn.it}),\\
Mohamed Elhoseiny$^{2}$
(\href{mailto:mohamed.elhoseiny@kaust.edu.sa}{mohamed.elhoseiny@kaust.edu.sa})

\noindent\textit{\textbf{Affiliations: }}\\
$^1$ Fudan University, China\\
$^2$ KAUST, Saudi Arabia\\
$^3$ MBZUAI, United Arab Emirates\\
$^4$ Tsinghua University, China\\
$^5$ University of W\"urzburg, Germany\\
$^6$ The Hong Kong Polytechnic University, China\\
$^7$ University of Trento, Italy\\

\subsection*{FDUROILab\_Lenovo}
\noindent\textit{\textbf{Title: }} FDUROILab\_Lenovo\\
\noindent\textit{\textbf{Members: }} \\
Lingyi Hong$^{1}$ (\href{mailto:lyhong22@m.fudan.edu.cn}{lyhong22@m.fudan.edu.cn}),\\
Mingxi Cheng$^{1}$ (\href{mailto:mxchen24@m.fudan.edu.cn}{mxchen24@m.fudan.edu.cn}),\\
Xingqi He$^{1}$,\\
Runze Li$^{2}$ (\href{mailto:lirz7@lenovo.com}{lirz7@lenovo.com}),\\
Xingdong Sheng$^{2}$ (\href{mailto:shengxd1@lenovo.com}{shengxd1@lenovo.com}),\\
Wenqiang Zhang$^{1,3}$ (\href{mailto:wqzhang@fudan.edu.cn}{wqzhang@fudan.edu.cn})\\
\noindent\textit{\textbf{Affiliations: }} \\
$^1$Shanghai Key Lab of Intelligent Information Processing, School of Computer Science, Fudan University, China\\
$^2$Lenovo Research\\
$^3$College of Intelligent Robotics and Advanced Manufacturing, Fudan University, China
\subsection*{CDiscover}
\noindent\textit{\textbf{Title: }} GiPL-Grounding: Generative augmented iterative Pseudo-Labeling for Grounding\\
\noindent\textit{\textbf{Members: }} \\
Jiacong Liu$^{1}$ (\href{mailto:m202574174@hust.edu.cn}{m202574174@hust.edu.cn}),\\
Shu Luo$^{1}$ (\href{mailto:luoshu_hust@hust.edu.cn}{luoshu\_hust@hust.edu.cn}),\\
Yikai Qin$^{1}$ (\href{mailto:yikaiq@hust.edu.cn}{yikaiq@hust.edu.cn}),\\
Yaze Zhao$^{1}$ (\href{mailto:zyaz@hust.edu.cn}{zyaz@hust.edu.cn}),\\
Yongwei Jiang$^{1}$ (\href{mailto:jiangyongwei@hust.edu.cn}{jiangyongwei@hust.edu.cn}),\\
Yixiong Zou$^{1}$ (\href{mailto:yixiongz@hust.edu.cn}{yixiongz@hust.edu.cn})\\
\noindent\textit{\textbf{Affiliations: }} \\
$^1$Huazhong University of Science and Technology
\subsection*{NJUST-KMG}
\noindent\textit{\textbf{Title: }} ASTER: A Hybrid Teacher-Guided Adaptation Framework for CD-FSOD\\
\noindent\textit{\textbf{Members: }} \\
Zhe Zhang$^{1\dagger}$ (\href{mailto:zhe.zhang@njust.edu.cn}{zhe.zhang@njust.edu.cn}),\\
Yang Yang$^{1}$ (\href{mailto:yyang@njust.edu.cn}{yyang@njust.edu.cn})\\
\noindent\textit{\textbf{Affiliations: }} \\
$^1$Nanjing University of Science and Technology
\subsection*{earth-insights}
\noindent\textit{\textbf{Title:} earth-insights}\\
\noindent\textit{\textbf{Members: }} \\
Kaiyu Li$^1$ (\href{mailto:likyoo.ai@gmail.com}{likyoo.ai@gmail.com}),\\
Bowen Fu$^1$ (\href{mailto:happybug@stu.xjtu.edu.cn}{happybug@stu.xjtu.edu.cn}),\\
Zixuan Jiang$^1$ (\href{mailto:andrewjiang@stu.xjtu.edu.cn}{andrewjiang@stu.xjtu.edu.cn}),\\
Ke Li$^2$ (\href{mailto:like0413@stu.xidian.edu.cn}{like0413@stu.xidian.edu.cn}),\\
Hui Qiao$^3$ (\href{mailto:qiaoh@chinatelecom.cn}{qiaoh@chinatelecom.cn}),\\
Xiangyong Cao$^1$ (\href{mailto:caoxiangyong@mail.xjtu.edu.cn}{caoxiangyong@mail.xjtu.edu.cn})\\
\noindent\textit{\textbf{Affiliations:}} \\
$^1$Xi’an Jiaotong University\\
$^2$Xidian University\\
$^3$China Telecom Shaanxi Branch
\subsection*{Intellindust AI Lab}
\noindent\textit{\textbf{Title: }} ZAP: Boosting Few-shot Object Detection with Auto-selected Zero-shot Pseudo Labels\\
\noindent\textit{\textbf{Members: }} \\
Xuanlong Yu$^1$ (\href{mailto:yuxuanlong@intellindust.com}{yuxuanlong@intellindust.com}),\\
Youyang Sha$^1$ (\href{mailto:shayouyang@intellindust.com}{shayouyang@intellindust.com}),\\
Longfei Liu$^1$ (\href{mailto:liulongfei@intellindust.com}{liulongfei@intellindust.com}),\\
Di Yang$^2$ (\href{mailto:di.yang@ustc.edu.cn}{di.yang@ustc.edu.cn}),\\
Xi Shen$^1$ (\href{mailto:shenxi@intellindust.com}{shenxi@intellindust.com})\\
\noindent\textit{\textbf{Affiliations: }} \\
$^1$Intellindust AI Lab\\
$^2$Suzhou Institute for Advanced Research, USTC
\subsection*{SAIDA}
\noindent\textit{\textbf{Title: }} Synthetic-Augmented Iterative Domain Adaptation\\
\noindent\textit{\textbf{Members: }} \\
Kyeongryeol Go,\\
Taewoong Jang\\
\noindent\textit{\textbf{Affiliations: }} \\
Superb AI
\clearpage
\subsection*{KLETech-CEVI}

\noindent\textit{\textbf{Title: }} Zero-Shot Transfer with GLIP and Multi-Scale Test-Time Augmentation\\

\noindent\textit{\textbf{Members: }} \\
Saiprasad Meesiyawar$^{3}$ (\href{mailto:saiprasad@cevi.co.in}{saiprasad@cevi.co.in}),\\
Ravi Kirasur$^{1,3}$ (\href{mailto:01fe23bcs233@kletech.ac.in}{01fe23bcs233@kletech.ac.in}),\\
Rakshita Kulkarni$^{1,3}$ (\href{mailto:01fe23bcs155@kletech.ac.in}{01fe23bcs155@kletech.ac.in}),\\
Bhoomi Deshpande$^{1,3}$ (\href{mailto:01fe23bcs145@kletech.ac.in}{01fe23bcs145@kletech.ac.in}),\\
Harsh Patil$^{1,3}$ (\href{mailto:01fe23bcs013@kletech.ac.in}{01fe23bcs013@kletech.ac.in}),\\
Uma Mudenagudi$^{2,3}$ (\href{mailto:uma@kletech.ac.in}{uma@kletech.ac.in})\\

\noindent\textit{\textbf{Affiliations: }} \\
$^1$School of Computer Science and Engineering, KLE Technological University, Hubballi, India\\
$^2$Department of Electronics and Communication Engineering, KLE Technological University, Hubballi, India\\
$^3$Center for Visual Intelligence (CEVI), Hubballi, India
\subsection*{Manifold}
\noindent\textit{\textbf{Title: }} Multimodal Prompt-Driven Diffusion Augmentation (MPDA)\\
\noindent\textit{\textbf{Members: }} \\
Shuming Hu (\href{mailto:hsm123@nudt.edu.cn}{hsm123@nudt.edu.cn}),\\
Chao Chen (\href{mailto:cc_19@nudt.edu.cn}{cc\_19@nudt.edu.cn}),\\
Tao Wang (\href{mailto:wtt977193@163.com}{wtt977193@163.com})\\
\noindent\textit{\textbf{Affiliations: }} \\
National University of Defence Technology, China
  \subsection*{QiFans}
  \noindent\textit{\textbf{Title: }} GDino-FT: Domain-Adaptive GroundingDINO with Prompt Engineering and Few-Shot Fine-Tuning\\
  \noindent\textit{\textbf{Members: }} \\
  Wei Zhou$^{1}$ (\href{mailto:weichow@u.nus.edu}{weichow@u.nus.edu})\\
  Qi Xu$^{2}$ (\href{mailto:txxqsh@gmail.com}{txxqsh@gmail.com})\\
  \noindent\textit{\textbf{Affiliations: }} \\
  $^{1}$National University of Singapore \\
  $^{2}$Shanghai Jiao Tong University
\subsection*{AIRCAS MILab}
\noindent\textit{\textbf{Title: }} Semantic-Filtered Domain-RAG and Auto-Tuned GroundingDINO\\
\noindent\textit{\textbf{Members: }} \\
Zhenzhao Xing (\href{mailto:xingzhenzhao25@mails.ucas.ac.cn}{xingzhenzhao25@mails.ucas.ac.cn}),\\
Dandan Zhao (\href{mailto:zhaodandan@aitech.edu.cn}{zhaodandan@aitech.edu.cn}),\\
Hanzhe Xia (\href{mailto:xiahanzhe25@163.com}{xiahanzhe25@163.com}),\\
Dongdong Lu (\href{mailto:ludd@aircas.ac.cn}{ludd@aircas.ac.cn}),\\
Zhe Zhang (\href{mailto:zhangzhe01@aircas.ac.cn}{zhangzhe01@aircas.ac.cn})\\
\noindent\textit{\textbf{Affiliations: }} \\
AIRCAS MILab
\subsection*{J\_G\_team}
\noindent\textit{\textbf{Title: }} Negative Prompting for Few-Shot Object Detection\\
\noindent\textit{\textbf{Members: }} \\
Jingru Wang (\href{mailto:wjr19950102@163.com}{wjr19950102@163.com}),\\
Guangwei Huang (\href{mailto:hgw61638181@gmail.com}{hgw61638181@gmail.com})\\
\noindent\textit{\textbf{Affiliations: }} \\
Free researcher, China
\subsection*{NTR}
\noindent\textit{\textbf{Title: }} NTR\\
\noindent\textit{\textbf{Members: }} \\
Jiachen Tu (\href{mailto:jtu9@illinois.edu}{jtu9@illinois.edu}),\\
Yaokun Shi (\href{mailto:yaokuns2@illinois.edu}{yaokuns2@illinois.edu}),\\
Guoyi Xu (\href{mailto:ericx3@illinois.edu}{ericx3@illinois.edu}),\\
Yaoxin Jiang (\href{mailto:yaoxinj2@illinois.edu}{yaoxinj2@illinois.edu}),\\
Jiajia Liu (\href{mailto:ciciliu2@illinois.edu}{ciciliu2@illinois.edu})\\
\noindent\textit{\textbf{Affiliations: }} \\
University of Illinois Urbana-Champaign
\subsection*{WRC}

\noindent\textit{\textbf{Title: }} MQ-Det\\
\noindent\textit{\textbf{Members: }} \\
Liwei Zhou (\href{mailto:zhouliwei123@nudt.edu.cn}{zhouliwei123@nudt.edu.cn}),\\
Bei Dou (\href{mailto:doubei@nudt.edu.cn}{doubei@nudt.edu.cn}),\\
Tao Wu (\href{mailto:wutao@nudt.edu.cn}{wutao@nudt.edu.cn})\\
\noindent\textit{\textbf{Affiliations: }} \\
College of Intelligence Science and Technology, National University of Defense Technology, China
\subsection*{NUDT-RSIP}

\noindent\textit{\textbf{Title: }} NUDT-RSIP\\
\noindent\textit{\textbf{Members: }} \\
Zekang Fan$^{1}$ (\href{mailto:fzk1055279850@qq.com}{fzk1055279850@163.com}),\\
Junjie Liu$^{1}$ (\href{mailto:junjieliucst@163.com}{junjieliucst@163.com})\\
\noindent\textit{\textbf{Affiliations: }} \\
$^1$National University of Defense Technology, China
\subsection*{French Borelli}

\noindent\textit{\textbf{Title: }} Triple-Tower\\
\noindent\textit{\textbf{Members: }} \\
Adhémar de Senneville (\href{mailto:adhemar.senneville@gmail.com}{adhemar.senneville@gmail.com}),\\
Flavien Armangeon (\href{mailto:flavien.armangeon@gmail.com}{flavien.armangeon@gmail.com})\\
\noindent\textit{\textbf{Affiliations: }} \\
Centre Borelli, ENS Paris-Saclay, 4 avenue des Sciences, Gif-sur-Yvette, France

\clearpage
\subsection*{FewShotEverything}
\noindent\textit{\textbf{Title:}} AIPR: Data Augmentation and Iterative Pseudo-labeling with Prototype Refinement for Cross-Domain Few-Shot Object Detection\\
\noindent\textit{\textbf{Members: }} \\
Yazhe Lyu\textsuperscript{1} (\href{yazhelv@hust.edu.cn}{yazhelv@hust.edu.cn})\\
Zhimeng Xin\textsuperscript{2} (\href{zhimengxin15@gmail.com}{zhimengxin15@gmail.com})\\
Zijian Zhuang\textsuperscript{1} (\href{zhuangzijian@hust.edu.cn}{zhuangzijian@hust.edu.cn})\\
\noindent\textit{\textbf{Affiliations: }} \\ 
\textsuperscript{1} School of Computer Science and Technology, Huazhong University of Science and Technology \\
\textsuperscript{2} School of Cyber Science and Engineering, Huazhong University of Science and Technology \\
\subsection*{Fusion-Few}
\noindent\textit{\textbf{Title: }} FusionFormer\\
\noindent\textit{\textbf{Members: }} \\
Hongchun Zhu (\href{mailto:vice2city@qq.com}{vice2city@qq.com})\\
\noindent\textit{\textbf{Affiliations: }} \\
University of Electronic Science and Technology of China, China.
\subsection*{nudt\_0110Dplter}
\noindent\textit{\textbf{Title: }} MSPF-ETS\\
\noindent\textit{\textbf{Members: }} \\
Wang Li (Team Leader) (\href{mailto:483617013@qq.com}{483617013@qq.com}),\\
Qisheng Xu (\href{mailto:qishengxu@nudt.edu.cn}{qishengxu@nudt.edu.cn}),\\
Changjian Wang (\href{mailto:wangcj@nudt.edu.cn}{wangcj@nudt.edu.cn}),\\
Kele Xu (\href{mailto:xukelele@nudt.edu.cn}{xukelele@nudt.edu.cn}),\\
Hui Geng (\href{mailto:gengh666666@163.com}{gengh666666@163.com}),\\
Xuyao Deng (\href{mailto:dengxuyao@nudt.edu.cn}{dengxuyao@nudt.edu.cn})\\
\noindent\textit{\textbf{Affiliations: }} \\
National University of Defense Technology

\clearpage
{
    \small
    \bibliographystyle{ieeenat_fullname}
    \bibliography{main}
}

\end{document}